\documentclass[journal]{IEEEtran}  
\ifCLASSINFOpdf
\else
\fi
\usepackage{amsmath,graphicx,cite}
\usepackage{subfig}
\usepackage{widetext}
\usepackage{color}
\usepackage{epstopdf}

\usepackage{tikz}
\usetikzlibrary{chains,shapes.multipart}
\usetikzlibrary{shapes,calc,fit}
\usetikzlibrary{automata,positioning}

\usepackage{algcompatible}

\usepackage[ruled]{algorithm2e}
\usepackage{algpseudocode}
\makeatletter
\renewcommand{\@algocf@capt@plain}{above}
\makeatother

\begin{document}
%
\title{Long-term Correlation Tracking using Multi-layer Hybrid Features in Sparse and Dense Environments}
%
%
%

\author{Nathanael~L.~Baisa,~\IEEEmembership{Member,~IEEE,}
        Deepayan~Bhowmik,~\IEEEmembership{Member,~IEEE,}
        and~Andrew~Wallace,~\IEEEmembership{Fellow,~IET,}
\thanks{N. L. Baisa and A. Wallace are with the Department of Electrical, Electronic and Computer Engineering, Heriot Watt University, Edinburgh EH14 4AS, United Kingdom. (e-mail: \{nb30, a.m.wallace\}@hw.ac.uk). D. Bhowmik is with the Department of Computing, Sheffield Hallam University, Sheffield S1 1WB, United Kingdom.(e-mail: deepayan.bhowmik@shu.ac.uk)}}
\maketitle

\begin{abstract}

Tracking a target of interest in both sparse and crowded environments is a challenging problem, not yet successfully addressed in the literature. In this paper, we propose a new long-term visual tracking algorithm, learning discriminative correlation filters and using an online classifier, to track a target of interest in both sparse and crowded video sequences. First, we learn a translation correlation filter using a multi-layer hybrid of convolutional neural networks (CNN) and traditional hand-crafted features. We combine advantages of both the lower convolutional layer which retains more spatial details for precise localization and the higher convolutional layer which encodes semantic information for handling appearance variations, and then integrate these with histogram of oriented gradients (HOG) and color-naming traditional features. Second, we include a re-detection module for overcoming tracking failures due to long-term occlusions by training an incremental (online) SVM on the most confident frames using hand-engineered features. This re-detection module is activated only when the correlation response of the object is below some pre-defined threshold. This generates high score detection proposals which are temporally filtered using a Gaussian mixture probability hypothesis density (GM-PHD) filter to find the detection proposal with the maximum weight as the target state estimate by removing the other detection proposals as clutter. Finally, we learn a scale correlation filter for estimating the scale of a target by constructing a target pyramid around the estimated or re-detected position using the HOG features. We carry out extensive experiments on both sparse and dense data sets which show that our method significantly outperforms state-of-the-art methods. 

\end{abstract}


%
\IEEEpeerreviewmaketitle

\section{Introduction}

Visual target tracking is one of the most important and active research areas in computer vision with a wide range of applications like surveillance, robotics and human-computer interaction (HCI). Although it has been studied extensively during past decades as recently surveyed in~\cite{SmeChuCuc14}\cite{YanShaZhe11}, object tracking is still a difficult problem due to many challenges that cause significant appearance changes of targets such as varying illumination, occlusion, pose variations, deformation, abrupt motion, background clutter, and high target densities (in crowded environments). 
Robust representation of target appearance is important to overcome these challenges.

Recently, convolutional neural network (CNN) features have demonstrated outstanding results on various recognition tasks~\cite{GirDonDar14,SimZis15}. Motivated by this, a few deep learning based trackers~\cite{WanYeu13,SeuTacSuh15} have been developed. In addition, discriminative correlation filter-based trackers have achieved state-of-the-art results as surveyed in~\cite{CheHonTao15} in terms of both efficiency and robustness due to three reasons. First, efficient correlation operations are performed by replacing exhausted circular convolutions with element-wise multiplications in the frequency domain which can be computed using the fast Fourier transform (FFT) with very high speed. Second, thousands of negative samples around the target's environment can be efficiently incorporated through circular-shifting with the help of a circulant matrix. Third, training samples are regressed to soft labels of a Gaussian function (Gaussian-weighted labels) instead of binary labels alleviating sampling ambiguity. In fact, regression with class labels can be seen as classification. However, correlation filter-based trackers are susceptible to long-term occlusions.

In addition, the Gaussian mixture probability hypothesis density (GM-PHD) filter~\cite{VoMa06} has an in-built capability of removing clutter while filtering targets with very efficient speed without the need for explicit data association. Though this filter is designed for multi-target filtering, it is even preferable for single target filtering in scenes with challenging background clutter as well as clutter that comes from other targets not of current concern. This filtering approach is flexible, for instance, it has been extended for multiple targets of different types in~\cite{BaiWal17}\cite{NatAnd17}.

In this work, we mainly focus on long-term tracking of a target of interest in sparse as well as crowded environments where the unknown target is initialized by a bounding box and then tracked in subsequent frames. Without making any constraint on the video scene, we develop a novel long-term online tracking algorithm that can close the research gap between sparse and crowded scenes tracking problems using the advantages of the correlation filters, a hybrid of multi-layer CNN and hand-crafted features, an incremental (online) support vector machine (SVM) classifier and a Gaussian mixture probability hypothesis density (GM-PHD) filter. To the best of our knowledge, nobody has adopted this approach.

The main contributions of this paper are as follows:
\begin{enumerate}
  \item 
      We integrate a hybrid of multi-layer CNN and traditional hand-crafted features for learning a translation correlation filter for estimating the target position in the next frame by extending a ridge regression for multi-layer features.
  \item 
      We include a re-detection module to re-initialize the tracker in case of tracking failures due to long-term occlusions by learning an incremental SVM from the most confident frames using hand-crafted features to generate high score detection proposals.
  \item 
      We incorporate a GM-PHD filter to temporally filter detection proposals generated from the learned online SVM to find the detection proposal with the maximum weight as the target position estimate by removing the other detection proposals as clutter.
  \item We learn a scale correlation filter by constructing a target pyramid at the estimated or re-detected position using HOG features to estimate the scale of the detected target.
\end{enumerate}

We presented a preliminary idea of this work in~\cite{NatDeeAnd17}. In this work, we make more elaborate descriptions of our algorithm. Besides, we include a scale estimation at the estimated target position as well as an extended experiment on a large-scale online object tracking benchmark (OOTB) in addition to the PETS 2009 data sets.

The rest of this paper is organized as follows. In section~\ref{sec:RelatedWork}, related work is discussed. An overview of our algorithm and the proposed algorithm in detail are described in sections~\ref{Sec:OverviewAlg} and~\ref{Sec:ProposedAlg}, respectively. In section~\ref{Sec:Implementation}, the implementation details with parameter settings is briefly discussed. The experimental results are analyzed and compared in section~\ref{Sec:ExperimentalResults}. The main conclusions and suggestions for future work are summarized in section~\ref{Sec:Conclusion}.

\section{Related Work} \label{sec:RelatedWork}

Various visual tracking algorithms have been proposed over the past decades to cope with tracking challenges, and they can be categorized into two types depending on the learning strategies: \textit{generative} and \textit{discriminative} methods. \textit{Generative} methods describe the target appearances using generative models and search for target regions that best-fit the models i.e. search for the best-matching windows (patches). Various generative target appearance modelling algorithms have been proposed such as online density estimation~\cite{HanComZhu08}, sparse representation~\cite{ZhaGhaLiu12,JiaLuYan12}, and incremental subspace learning~\cite{RosLimLin08}. On the other hand, \textit{discriminative} methods build a model that distinguishes the target from the background. These algorithms typically learn classifiers based on online boosting~\cite{GraLeiBis08}, multiple instance learning~\cite{BabYanBel11}, P-N learning~\cite{ZdeKryJir12}, transfer learning~\cite{JinHaiWei14}, structured output SVMs~\cite{HarSafTor11} and combining multiple classifiers with different learning rates~\cite{ZhaMaScl14}. Background information is important for effective tracking as explored in~\cite{WuLimYan13}\cite{WuLimYan15} which means that more competing approaches are discriminative methods~\cite{Tom05} though hybrid generative and discriminative models can also be used~\cite{DinYuMed14}\cite{ZhoLuYan12}. However, sampling ambiguity is one of the big problems in discriminative tracking methods which results in drifting. Recently, correlation filters~\cite{HenCasMar12,HenCasMar14,DanHagSha14} have been introduced for online target tracking that can alleviate this sampling ambiguity. Previously, the large training data required to train correlation filters prevented them from application to online visual tracking though correlation filters are effective for localization tasks. However, recently all the circular-shifted versions of input features have been considered with the help of a circulant matrix producing a large number training samples~\cite{HenCasMar12,HenCasMar14}.

There are many strong sides of correlation methods such as inherent parallelism, shift (translation) invariance, noise robustness, and high discrimination ability~\cite{WanAlfBro17}. Both digital and optical correlators are discussed in detail in~\cite{AlfBro15} though more emphasis is given to optical correlators. Performance optimization of the correlation filters by pre-processing the input target image was introduced in~\cite{BouElbAlfBroFak16}.  Recent research trends of correlation filters for various applications with more emphasis on face recognition (and object tracking) is given in~\cite{WanAlfBro17}. Due to the effectiveness of the correlation methods, they have been successfully applied to many domains such as swimmer tracking~\cite{BenNapAlfVerHel17}, pose invariant face recognition~\cite{NapAlf17}, road sign identification for advanced driver assistance~\cite{OueAlfDesBro17}, etc. Some types of correlation filters are sensitive to challenges such as rotation, illumination changes, occlusion, etc. For instance, the Phase-Only Filter (POF) is sensitive to changes in rotation, illumination changes, occlusion, scale and noise contained in targets of interest~\cite{BouElbAlfBroFak16} though it can give very narrow correlation peaks (good localization); a pre-processing step was used to make it invariant to illumination in~\cite{NapAlf17}. Recent correlation filters such as KCF~\cite{HenCasMar14} are more suitable for online tracking by generating a large number of training samples from input features using a circulant matrix and are more robust to the tracking challenges such as rotation, illumination changes, partial occlusion, deformation, fast motion, etc (as shown on its results section in~\cite{HenCasMar14}) than its previous counterparts~\cite{WanAlfBro17}. Using CNN features has even improved the online tracking results as shown in~\cite{MaHunYan15} against these tracking challenges, however, log-term occlusion is still a problem in correlation filter-based tracking approaches.

There are three tracking scenarios that are important to consider: short-term tracking, long-term tracking, and tracking in a crowded scene. If objects are visible over the whole course of the sequences, short-term model-free tracking algorithms are sufficient to track a single object without applying a pre-trained model of target appearance. There are many short-term tracking algorithms developed in the literature~\cite{SmeChuCuc14}\cite{CheHonTao15} such as online density estimation~\cite{HanComZhu08}, context-learning~\cite{KaiLeiQin14}, scale estimation~\cite{DanHagSha14}, and using features from multiple CNN layers~\cite{MaHunYan15,WanOuyWan15}. However, these short-term tracking algorithms can not re-initialize the trackers once they fail due to long-term occlusions and confusions from background clutter.

Long-term tracking algorithms are important in video streams that are of indefinite length and have long-term occlusions. A Tracking-Learning-Detection (TLD) algorithm has been developed in~\cite{ZdeKryJir12} which explicitly decomposes the long-term tracking task into tracking, learning and detection. In this case, the tracker tracks the target from frame to frame and provides training data for the detector which re-initializes the tracker when it fails. The learning component estimates the detector's errors and then updates it for correction in the future. This algorithm works well in very sparse video (video sequences with few targets) but is sensitive to background clutter. Long-term correlation tracking (LCT), developed in~\cite{MaYanZha15}, learns three different discriminative correlation filters: translation, appearance and scale correlation filters using hand-crafted features. Even though it includes a re-detection module by learning the random ferns classifier online for re-initializing a tracker in case of tracking failures, it is not robust to long-term occlusions and background clutter. Multi-domain network (MDNet)~\cite{HyeBoh15} pre-trains a CNN network composed of shared layers and multiple domain-specific layers using a large set of videos to get generic target representations in the shared layers. This proposed network has separate branches of domain-specific layers for binary classification to identify the target in each domain. However, when applied to fundamentally different videos other than the related videos on which it was trained, it gives poorer results.

Tracking of a target in a crowded scene is very challenging due to long-term occlusion, many targets with appearance variation and high clutter. Person detection and tracking in crowds is formulated as a joint energy minimization problem by combining crowd density estimation and localization of individual person in~\cite{RodSivLap11}. Although this approach does not require manual initialization, it has low performance for tracking a generic target of interest as it was mainly developed for tracking human heads. The method developed in~\cite{KraNis12} trained Hidden Markov Models (HMMs) on motion patterns within a scene to capture spatial and temporal variations of motion in the crowd which is used for tracking individuals. However, this approach is limited to a crowd with structured pattern i.e. it needs some prior knowledge about the scene. The algorithm developed in~\cite{HarNolMub13} used visual information (prominence) and spatial context (influence from neighbours) to develop online tracking in crowded scene without using any prior knowledge about the scene, unlike the method in~\cite{KraNis12} which uses some training data from the past as well as the future. This algorithm performs well in highly crowded scenes but has low performance in a less crowded scenes as influence from neighbours (spatial context) decreases.

In conclusion, although there are many effective algorithms that handle appearance variation, occlusion and high clutter in short and long-term video sequences, no single approach is wholly effective in all scenarios. Our proposed tracking algorithm tracks a target of interest in both sparse and dense environments without using any constraint from the video scene using correlation filters, sophisticated features and re-detection scheme particularly robust to sparse as well as highly occluded and cluttered dense scenes.

\section{Overview of Our Algorithm}  \label{Sec:OverviewAlg}


We develop a novel long-term online tracking algorithm that can be applied to both sparse and dense environments by learning correlation filters using a hybrid of multi-layer CNN and hand-crafted features as well as including a re-detection module using an incremental SVM and GM-PHD filter.

Accordingly, to develop an online long-term tracking algorithm robust to appearance variations in both sparse and crowded scenes, we learn two different correlation filters: a translation correlation filter ($\mathbf{w}_t$) and a scale correlation filter ($\mathbf{w}_s$). A translation correlation filter is learned using a hybrid of multi-layer CNN features from VGG-Net~\cite{SimZis15} and robust traditional hand-crafted features.

For the CNN part, we combine features from both a lower convolutional layer which retains more spatial details for precise localization and a higher convolutional layer which encodes semantic information for handling appearance variations. This makes layer 1, layer 2 and layer 3 in multi-layer features with multiple channels (512, 512 and 256 dimensions) in each layer, respectively. Since the spatial resolution of the extracted features gradually reduces with the increase of the depth of CNN layers due to pooling operators, it is crucial to resize each feature map to a fixed size using bilinear interpolation.

For the traditional features part, we use a histogram of oriented gradients (HOG), in particular Felzenszwalb's variant~\cite{FelGirMcA10} and color-naming~\cite{VanSchVer09} features for capturing image gradients and color information, respectively. These integrated traditional features were used for object detection in~\cite{KhaAnwWei12}\cite{LiZhu15} giving promising results. Color-naming is the linguistic color label assigned by human to describe the color, hence, the mapping method in~\cite{VanSchVer09} is employed to convert the RGB space into the color name space which is an 11 dimensional color representation providing the perception of a target color. By aligning the feature size of the HOG variant with 31 dimensions and color-naming with 11 dimensions, they are integrated to make 42 dimensional features which make a 4th layer in our hybrid multi-layer features.

\begin{figure*}[!htb]
\begin{center}
   \includegraphics[width=1.0\linewidth]{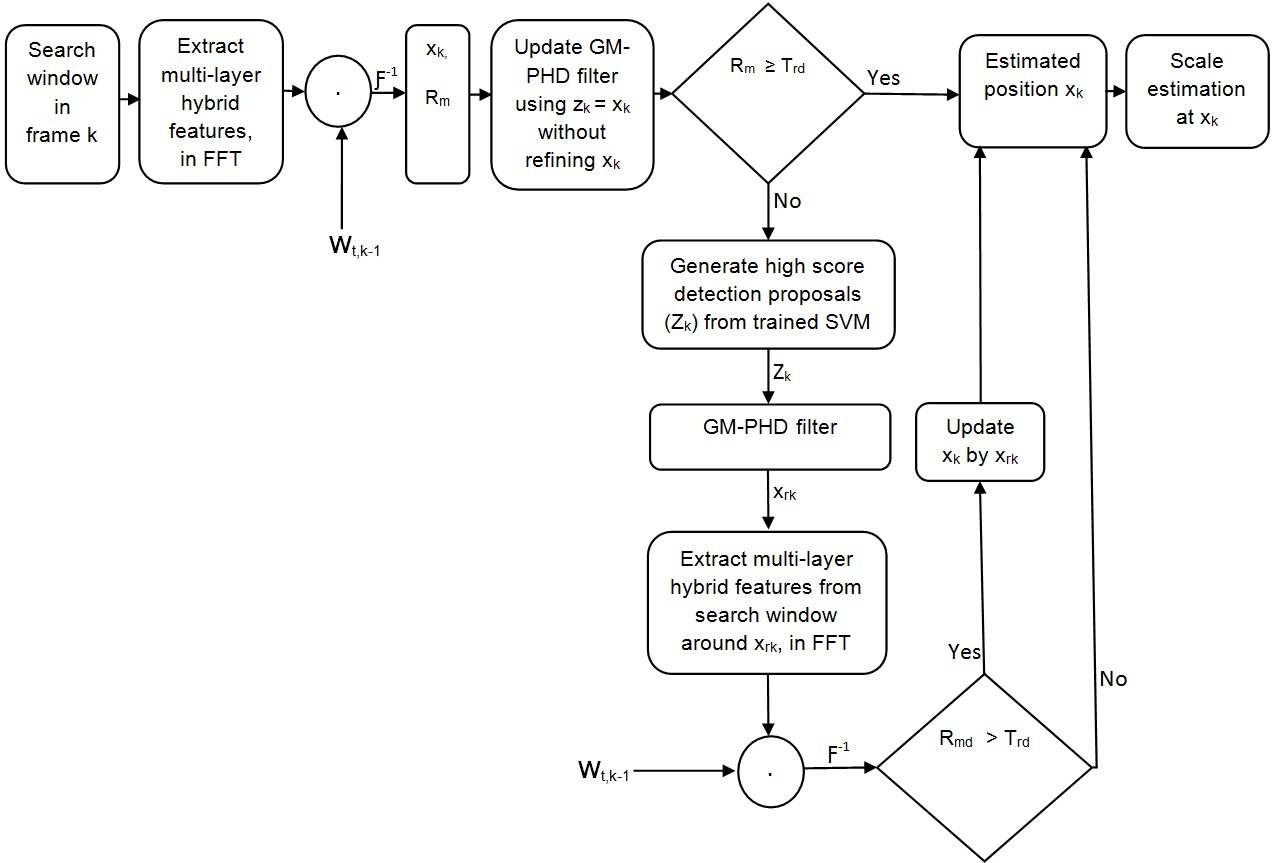} \\
\end{center}
   \caption{\small{The flowchart of the proposed algorithm. It consists of three main parts: translation estimation, re-detection and scale estimation. Given a search window, we extract multi-layer hybrid features (in the frequency domain) and then estimate target position ($\mathbf{x}_{k}$) using a translation correlation filter ($\mathbf{w}_t$). This estimated position ($\mathbf{x}_{k}$) is used as a measurement ($\mathbf{z}_{k}$) for updating the GM-PHD filter without refining $\mathbf{x}_{k}$, just to update its weight for later use during re-detection. Re-detection is activated if the maximum of the response map ($R_m$) becomes below the pre-defined threshold ($T_{rd}$). Then, we generate high score detection proposals ($Z_k$) which are filtered by the GM-PHD filter to estimate the detection with maximum weight as target position ($\mathbf{x}_{rk}$) removing the others as clutter. If the response map around $\mathbf{x}_{rk}$ ($R_{md}$) is greater than $T_{rd}$, the target position $\mathbf{x}_{k}$ is updated by the re-detected position $\mathbf{x}_{rk}$. Finally, we estimate the scale of the target by constructing a target pyramid at the estimated position and use the scale correlation filter ($\mathbf{w}_s$) to find the scale at which the maximum response map is obtained. Note that in frame 1, we only train correlation filters and the SVM classifier using the initialized target; no detection is performed.}} 
\label{fig:BlockDiagramLCMHT}
\end{figure*}

For scale estimation, we learn a scale correlation filter using only HOG features, in particular Felzenszwalb's variant~\cite{FelGirMcA10}. Besides, we incorporate a re-detection module by learning an incremental SVM from the most confident frames determined by maximal value of correlation response map using HOG, LUV color and normalized gradient magnitude features for generating high-score detection proposals which are filtered using the GM-PHD filter to re-acquire the target in case of tracking failures. The flowchart of our method is given in Fig.~\ref{fig:BlockDiagramLCMHT} and the outline of our proposed algorithm is given in Algorithm~\ref{alg:ProposedTraAlg}.

\noindent

\section{Proposed Algorithm}  \label{Sec:ProposedAlg}

This section describes our proposed tracking algorithm which has four distinct functional parts: a) correlation filters formulated for multi-layer hybrid features, b) an online SVM detector developed for generating high score detection proposals, c) a GM-PHD filter for finding the detection proposal with maximum weight to re-initialize the tracker in case of tracking failures by removing the other detection proposals as clutter, and d) a scale estimation method for estimating the scale of a target by constructing image pyramid at the estimated target position.

\subsection{Correlation Filters for Multi-layer Features}\label{Subsec:CorrelationFilters}

To track a target using correlation filters, the appearance of the target should be modeled using a correlation filter $\mathbf{w}$ which can be trained on feature vector $\mathbf{x}$ of size $M \times N \times D$ extracted from an image patch where M, N, and D indicates the width, height and number of channels, respectively. This feature vector $\mathbf{x}$ can be extracted from multiple layers, for example in the case of CNN features and/or traditional hand-crafted features, therefore, we denote it as $\mathbf{x}^{(l)}$ to designate from which layer $l$ it is extracted. All the circular shifts of $\mathbf{x}^{(l)}$ along the $M$ and $N$ dimensions are considered as training examples where each circularly shifted sample $\mathbf{x}^{(l)}_{m,n}, m \in \{0,1,...,M-1\}, n\in \{0,1,...,N-1\}$ has a Gaussian function label $y^{(l)}(m,n)$ given by

\begin{equation}
     y^{(l)}(m,n) = e^{-\frac{(m-M/2)^{2} + (n-N/2)^{2}}{2\sigma^2}},
\label{eqn:GaussianLabel}
\end{equation}
\noindent where $\sigma$ is the kernel width. Hence, $y^{(l)}(m,n)$ is a soft label rather than a binary label. To learn the correlation filter $\mathbf{w}^{(l)}$ for layer $l$ with the same size as $\mathbf{x}^{(l)}$, we extend a ridge regression~\cite{RifYeoPog03}\cite{Mur12}, developed for a single-layer feature vector, to be used for multi-layer hybrid features with layer $l$ as

\begin{equation}
     \underset{\mathbf{w}^{(l)}}{\text{min}~} \sum_{m,n} |\Phi(\mathbf{x}^{(l)}). \mathbf{w}^{(l)} - y^{(l)}(m,n)|^{2}  + \lambda |\mathbf{w}^{(l)}|^{2},
\label{eqn:RidgeRegression}
\end{equation}
\noindent where $\Phi$ denotes the mapping to a kernel space and $\lambda$ is a regularization parameter ($\lambda \geq 0$). The solution $\mathbf{w}^{(l)}$ can be expressed as

\begin{equation}
     \mathbf{w}^{(l)} = \sum_{m,n} \mathbf{a}^{(l)}(m,n)\Phi(\mathbf{x}_{m,n}^{(l)}),
\label{eqn:CorrelationFilterW}
\end{equation}
\noindent This alternative representation makes the dual space $\mathbf{a}^{(l)}$ the variable under optimization instead of the primal space $\mathbf{w}^{(l)}$.

\textbf{Training phase:} The training phase is performed in the Fourier domain using the fast Fourier transform (FFT) to compute the coefficient $\mathbf{A}^{(l)}$ as

\begin{equation}
     \mathbf{A}^{(l)} = \mathcal{F}(\mathbf{a}^{(l)})  = \frac{\mathcal{F}(\mathbf{y}^{(l)})}{\mathcal{F}\big(\Phi(\mathbf{x}^{(l)}).\Phi(\mathbf{x}^{(l)})\big) + \lambda},
\label{eqn:CorrelationFilterA}
\end{equation}
\noindent where $\mathcal{F}$ denotes the FFT operator.

\textbf{Detection phase:} The detection phase is performed on the new frame given an image patch (search window) which is used as a temporal context i.e. the search window is larger than the target to provide some context. If feature vector $\mathbf{z}^{(l)}$ of size $M \times N \times D$ is extracted from this image patch, the response map $(\mathbf{r}^{(l)})$ is computed as
\begin{equation}
     \mathbf{r}^{(l)} = \mathcal{F}^{-1}\big(\tilde{\mathbf{A}}^{(l)} \odot \mathcal{F}(\Phi(\mathbf{z}^{(l)}).\Phi(\tilde{\mathbf{x}}^{(l)}))\big),
\label{eqn:CorrelationFilterR}
\end{equation}
\noindent where $\tilde{\mathbf{A}}^{(l)}$ and $\tilde{\mathbf{x}}^{(l)} =  \mathcal{F}^{-1}(\tilde{\mathbf{X}}^{(l)})$ denote the learned target appearance model for layer $l$, operator $\odot$ is the Hadamard (element-wise) product, and $\mathcal{F}^{-1}$ is the inverse FFT. Now, the response maps of all layers are summed according to their weight $\gamma(l)$ element-wise as

\begin{equation}
     \mathbf{r}(m,n) = \sum_l \gamma(l)\mathbf{r}^{(l)}(m,n),
\label{eqn:CorrelationFilterRC}
\end{equation}
\noindent

The new target position is estimated by finding the maximum value of $\mathbf{r}(m,n)$ as
\begin{equation}
     (\hat{m}, \hat{n}) = \underset{m, n}{\text{argmax}~} \mathbf{r}(m,n),
\label{eqn:EstimatedPosition}
\end{equation}
\noindent

\textbf{Model update:} The model is updated by training a new model at the new target position and then linearly interpolating the obtained values of the dual space coefficients $\mathbf{A}^{(l)}_k$ and the base data template $\mathbf{X}^{(l)}_k = \mathcal{F} (\mathbf{x}^{(l)}_k)$ with those from the previous frame to make the tracker more adaptive to target appearance variations.

\begin{subequations}
\begin{align}
     \tilde{\mathbf{X}}^{(l)}_k &= (1 - \eta)\tilde{\mathbf{X}}^{(l)}_{k-1} + \eta \mathbf{X}^{(l)}_k, \\
     \tilde{\mathbf{A}}^{(l)}_k &= (1 - \eta)\tilde{\mathbf{A}}^{(l)}_{k-1} + \eta \mathbf{A}^{(l)}_k,
\end{align}
\label{eqn:CorrelattionFilterModelUpdate}
\end{subequations}
\noindent where $k$ is the index of the current frame, and $\eta$ is the learning rate.

The mappings to the kernel space $(\Phi)$ used in Eq.~(\ref{eqn:CorrelationFilterA}) and Eq.~(\ref{eqn:CorrelationFilterR}) can be expressed using a kernel function as $K(\mathbf{x}^{(l)}_i,\mathbf{x}^{(l)}_j) = \Phi(\mathbf{x}^{(l)}_i).\Phi(\mathbf{x}^{(l)}_j) = \Phi(\mathbf{x}^{(l)}_i)^{T}\Phi(\mathbf{x}^{(l)}_j)$. If the computation is performed in the frequency domain, the normal transpose should be replaced by the Hermitian transpose i.e. $\Phi(\mathbf{X}^{(l)}_i)^{H} = (\Phi(\mathbf{X}^{(l)}_i)^*)^{T}$  where the star ($*$) denotes the complex conjugate.

Thus, for a linear kernel,
\begin{equation}
     K(\mathbf{x}^{(l)}_i,\mathbf{x}^{(l)}_j) = (\mathbf{x}^{(l)}_i)^{T}\mathbf{x}^{(l)}_j = \mathcal{F}^{-1}(\sum_d (\mathbf{X}^{(l)}_{i,d})^{*} \odot \mathbf{X}^{(l)}_{j,d}),
\label{eqn:LinearKernel} 
\end{equation}
\noindent where $\mathbf{X}^{(l)}_{i} =  \mathcal{F}(\mathbf{x}^{(l)}_{i})$. 

and for a Gaussian kernel,
\begin{equation}
\begin{array} {lll}
     K(\mathbf{x}^{(l)}_i,\mathbf{x}^{(l)}_j) =  \Phi(\mathbf{x}^{(l)}_i)^{T}\Phi(\mathbf{x}^{(l)}_j) = \exp{\big(-\frac{|\mathbf{x}^{(l)}_i - \mathbf{x}^{(l)}_j |^{2}}{\sigma^{2}} \big)} = \\ \exp{\bigg(-\frac{1}{\sigma^{2}} \big( \|\mathbf{x}^{(l)}_i \|^{2} +  \|\mathbf{x}^{(l)}_j \|^{2} - \mathcal{F}^{-1}(\sum_d (\mathbf{X}^{(l)}_{i,d})^{*} \odot \mathbf{X}^{(l)}_{j,d}) \big) \bigg)},
\label{eqn:GaussianKernel}
\end{array}
\end{equation}
\noindent
This formulation is generic for multiple channel features from multiple layers as in the case of our multi-layer hybrid features, i.e. where $\mathbf{X}^{(l)}_{i,d},~ d \in \{1,...,D\}, ~l \in \{1, ..., L\}$. This is an extended version of the one given in~\cite{HenCasMar14} that takes into account features from multiple layers. The linearity of the FFT allows us to simply sum the individual dot-products for each channel $d \in \{1,...,D\}$ in each layer $l \in \{1,...,L\}$.

\subsection{Online Detector}

We include a re-detection module, $D_r$, to generate high score detection proposals in case of tracking failures due to long-term occlusions. Instead of using a correlation filter to scan across the entire frame which is computationally expensive and less efficient, we learn an incremental (online) SVM~\cite{DieCau03} by generating a set of samples in the search window around the estimated target position from the most confident frames, and scan through the window when the re-detection is activated to generate high-score detection proposals. These most confident frames are determined by the maximum translation correlation response in the current frame i.e. if the maximum correlation response of an image patch is above the trained detector threshold ($T_{td}$), we generate samples around this image patch and train the detector. This detector is activated to generate high score detection proposals if the maximum of the correlation response becomes below the activated re-detection threshold ($T_{rd}$). We use HOG (particularly Felzenszwalb's variant~\cite{FelGirMcA10}), LUV color and normalized gradient magnitude features to train this online SVM classifier. We use different visual features for computational feasibility from the ones we use for learning the translation correlation filter since we can select the feature representation for each module independently~\cite{DanHagSha14,MaYanZha15}.

We want to update a weight vector $\mathbf{w}$ of the SVM provided a set of samples with associated labels, $\{(\acute{\mathbf{x}}_i, \acute{\mathbf{y}}_i)\}$, obtained from the current results. The label $\acute{\mathbf{y}}_i$ of a new example $\acute{\mathbf{x}}_i$ is given by

\begin{equation}
    \acute{\mathbf{y}}_i =
  \begin{cases}
    +1,       & \quad \text{if } IOU(\acute{\mathbf{x}}_i, \ddot{\mathbf{x}}_t) \geq \delta_p \\
    -1,  & \quad \text{if } IOU(\acute{\mathbf{x}}_i, \ddot{\mathbf{x}}_t) < \delta_n \\
  \end{cases}
\label{eqn:label1}
\end{equation}
\noindent where $IOU(.)$ is the intersection over union (overlap ratio) of a new example $\acute{\mathbf{x}}_i$ and the estimated target bounding box in the current most confident frame $\ddot{\mathbf{x}}_t$. The samples with the bounding box overlap ratios between the thresholds $\delta_n$ and $\delta_p$ are excluded from the training set for avoiding the drift problem.

SVM classifiers of the form $f(\mathbf{x}) = \mathbf{w}.\Phi(\mathbf{x}) + b$ are learned from the data $\{(\mathbf{x}_i, \mathbf{y}_i) \in \Re^{m} \times \{-1, +1\} \forall i \in \{1,...,N\}\}$ by minimizing

\begin{equation}
    \underset{\mathbf{w},b,\boldmath{\xi}}{\text{min}} ~~ \frac{1}{2} ||\mathbf{w}||^{2} + C \sum_{i=1}^{N} \xi_i^{p}
\label{eqn:SVM}
\end{equation}
\noindent for $p \in \{1, 2\}$ subject to the constraints
\begin{equation}
    y_i(\mathbf{w}.\Phi(\mathbf{x_i}) + b) \geq 1 - \xi_i, \xi_i \geq 0~ \forall i \in \{1, ..., N\}.
\label{eqn:SVMconstraints}
\end{equation}
\noindent Hinge loss ($p = 1$) is preferred due to its improved robustness to outliers over the quadratic loss ($p = 2$). Thus, the offline SVM learns a weight vector $\mathbf{w} = (w_1, w_2, ...., w_N)^{T}$ by solving this quadratic convex optimization problem (QP) which can be expressed in its dual form as

\begin{equation}
     \underset{0\leq a_i\leq C}{\text{min}} W =  \frac{1}{2} \sum_{i,j =1}^{N} a_iQ_{ij}a_j - \sum_{i=1}^{N} a_i + b \sum_{i=1}^{N} y_i a_i,
\label{eqn:ObjectiveFn}
\end{equation}
\noindent where $\{a_i\}$ are Lagrange multipliers, $b$ is bias, $C$ is regularization parameter, and $Q_{ij} = y_i y_j K(\mathbf{x_i},\mathbf{x_j})$. The kernel function $K(\mathbf{x_i}, \mathbf{x_j}) = \Phi(\mathbf{x_i}).\Phi(\mathbf{x_j})$ is used to implicitly map into a higher dimensional feature space and compute the dot product. It is not straightforward for conventional QP solvers to handle the optimization problem in Eq.~(\ref{eqn:ObjectiveFn}) for online tracking tasks as the training data are provided sequentially, not at once. Incremental SVM~\cite{DieCau03} is tailored for such cases which retains Karush-Kuhn-Tucker (KKT) conditions on all the existing examples while updating the model with a new example so that the exact solution at each increment of dataset can be guaranteed. KKT conditions are the first-order necessary conditions for the optimal unique solution of dual parameters $\{a, b\}$ which minimizes Eq.~(\ref{eqn:ObjectiveFn}) and are given by

\begin{equation}
    \frac{\partial W}{\partial a_i} = \sum_{j=1}^{N} Q_{ij} a_j + y_i b - 1 \begin {cases} & \quad \hspace{-7mm} > 0, \text{if } a_i = 0 \\
                                                                            & \quad \hspace{-7mm} = 0, \text{if } 0 \leq a_i \leq C \\
                                                                            & \quad \hspace{-7mm} < 0, \text{if } a_i = C,
                                                                    \end{cases}
\label{eqn:KKTconditions1}
\end{equation}
\noindent
\begin{equation}
      \frac{\partial W}{\partial b} = \sum_{j=1}^{N} y_j a_j = 0,
\label{eqn:KKTconditions2}
\end{equation}
\noindent Based on the partial derivative $m_i = \frac{\partial W}{\partial a_i}$ which is related to the margin of the i-th example, each training example can be categorized into three: $\mathcal{S}_1$ support vectors lying on the margin ($m_i = 0$), $\mathcal{S}_2$ support vectors lying inside the margin ($m_i < 0$), and the remaining $\mathcal{R}$ reserve vectors (non-support vectors). During incremental learning, new examples with $m_i \leq 0$ eventually become margin ($\mathcal{S}_1$) or error ($\mathcal{S}_2$) support vectors. However, the remaining new training examples become reserve vectors as they do not enter the solution so that the Lagrangian multipliers ($a_i$) are estimated while retaining the KKT conditions. Given the updated Lagrangian multipliers, the weight vector $\mathbf{w}$ is given by

\begin{equation}
      \mathbf{w} = \sum_{i\in \mathcal{S}_1 \cup \mathcal{S}_2} a_i y_i\Phi(\mathbf{x_i}),
\label{eqn:KKTw}
\end{equation}
\noindent It is important to keep only a fixed number of support vectors with the smallest margins for efficiency during online tracking.

Thus, using the trained incremental SVM, we generate high score detections as detection proposals during the re-detection stage. These are filtered using the GM-PHD filter to find the best possible detection that can re-initialize the tracker.

\subsection{Temporal Filtering using the GM-PHD Filter}

Once we generate high score detection proposals using the learned online SVM classifier during the re-detection stage, we need to find the most probable detection proposal for the target state (position) estimate by finding the detection proposal with the maximum weight using the GM-PHD filter~\cite{VoMa06}. Though the GM-PHD filter is designed for multi-target filtering with the assumptions of a linear Gaussian system, in our problem (re-detecting a target in cluttered scene), it is used for removing clutter that comes from background scene and other targets not of interest as it is equipped with such a capability. Besides, it provides motion information for the tracking algorithm. More importantly, using the GM-PHD filter to find the detection with the maximum weight from the generated high score detection proposals is more robust than relying only on the maximum score of the classifier.

The detected position of the target in each frame is filtered using the GM-PHD filter, but without re-fining the position states until the re-detection module is activated. This updates the weight of the GM-PHD filter corresponding to a target of interest giving sufficient prior information to be picked up during the re-detection stage among candidate high score detection proposals. If the re-detection module is activated (correlation response of the target becomes below a pre-defined threshold), we generate high score detection proposals (in this case 5) from the trained SVM classifier which are then filtered using the GM-PHD filter. The Gaussian component with the maximum weight is selected as the position estimate, and if the correlation response of this estimated position is greater than the pre-defined threshold, the estimated position of the target is re-fined.

The GM-PHD filter has two steps: prediction and update. Before stating these two steps, certain assumptions are needed: 1) each target follows a linear Gaussian model:

\begin{equation}
    y_{k|k-1}(x|\zeta) =  \mathcal{N}(x;F_{k-1}\zeta, Q_{k-1})
\label{eqn:linearState1}
\end{equation}
\noindent
\begin{equation}
    f_{k}(z|x) =  \mathcal{N}(z;H_{k} x, R_{k})
\label{eqn:linearObservation1}
\end{equation}
\noindent where $\mathcal{N}(.;m, P)$ denotes a Gaussian density with mean $m$ and covariance $P$; $F_{k-1}$ and $H_k$ are the state transition and measurement matrices, respectively. $Q_{k-1}$ and $R_k$ are the covariance matrices of the process and the measurement noises, respectively.
2) A current measurement driven birth intensity inspired by but not identical to~\cite{RisClaVoVo12} is introduced at each time step, removing the need for the prior knowledge (specification of birth intensities) or a random model, with a non-informative zero initial velocity. The intensity of the spontaneous birth RFS is a Gaussian mixture of the form

\begin{equation}
\begin{split}
     \gamma_{k}(x)  =  \sum_{v = 1}^{V_{\gamma,k}} w_{\gamma,k}^{(v)}\mathcal{N}(x; m_{\gamma,k}^{(v)}, P_{\gamma,k}^{(v)})
\label{eqn:PHDbirthassumption2}
\end{split}
\end{equation}
\noindent where $V_{\gamma,k}$ is the number of birth Gaussian components, $w_{\gamma,k}^{(v)}$ is the weight accompanying the Gaussian component $v$, $m_{\gamma,k}^{(v)}$ is the current measurement and zero initial velocity used as mean, and $P_{\gamma,k}^{(v)}$ is birth covariance for Gaussian component $v$. In our case, $V_{\gamma,k}$ equals to 1 unless in re-detection stage at which it becomes 5 as we generate 5 high score detection proposals to be filtered.

3) The survival and detection probabilities are independent of the target state: $p_{s,k}(x_k) = p_{s,k}$ and $p_{D,k}(x_k) = p_{D,k}$.

\textbf{Prediction:} It is assumed that the posterior intensity at time $k-1$ is a Gaussian mixture of the form

\begin{equation}
\begin{split}
     \mathcal{D}_{k-1}(x)  =  \sum_{v = 1}^{V_{k-1}} w_{k-1}^{(v)}\mathcal{N}(x; m_{k-1}^{(v)}, P_{k-1}^{(v)}),
\label{eqn:PHDposterior1k-1}
\end{split}
\end{equation}

\noindent where $V_{k-1}$ is the number of Gaussian components of $\mathcal{D}_{k-1}(x)$ and it equals to the number of Gaussian components after pruning and merging at the previous iteration. Under these assumptions, the predicted intensity at time $k$ is given by

\begin{equation}
    \mathcal{D}_{k|k-1}(x) = \mathcal{D}_{S,k|k}(x) + \gamma_{k}(x),
\label{eqn:PHDpredictionI1}
\end{equation}
\noindent where

\begin{equation}
\begin{array} {lll}  \mathcal{D}_{S,k|k-1}(x) =& p_{s,k} \sum_{v = 1}^{V_{k-1}} w_{k-1}^{(v)}\mathcal{N}(x;  m_{S,k|k-1}^{(v)},P_{S,k|k-1}^{(v)}), \nonumber
\end{array}
\label{eqn:PHDpredictionSurvival1}
\end{equation}
\noindent
\begin{equation}
 m_{S,k|k-1}^{(v)} = F_{k-1} m_{k-1}^{(v)},  \nonumber
\label{eqn:PHDpredictionSurvivalMean1}
\end{equation}
\noindent
\begin{equation}
 P_{S,k|k-1}^{(v)} = Q_{k-1} + F_{k-1} P_{k-1}^{(v)} F^T_{k-1},  \nonumber
\label{eqn:PHDpredictionSurvivalCov1}
\end{equation}
\noindent where $\gamma_k(x)$ is given by Eq.~(\ref{eqn:PHDbirthassumption2}).

Since $\mathcal{D}_{S,k|k-1}(x)$ and $\gamma_k(x)$ are Gaussian mixtures, $ \mathcal{D}_{k|k-1}(x)$ can be expressed as a Gaussian mixture of the form

\begin{equation}
\begin{split}
     \mathcal{D}_{k|k-1}(x)  =  \sum_{v = 1}^{V_{k|k-1}} w_{k|k-1}^{(v)}\mathcal{N}(x; m_{k|k-1}^{(v)},P_{k|k-1}^{(v)}),
\label{eqn:PHDpredictionki}
\end{split}
\end{equation}
\noindent where $w_{k|k-1}^{(v)}$ is the weight accompanying the predicted Gaussian component $v$, and $V_{k|k-1}$ is the number of predicted Gaussian components and it equals to the number of born targets (1 unless in case of re-detection at which it is 5) and the number of persistent components which are actually the number of Gaussian components after pruning and merging at the previous iteration.

\textbf{Update:} The posterior intensity (updated PHD) at time $k$  is also a Gaussian mixture and is given by
\begin{equation}
\begin{split}
     \mathcal{D}_{k|k}(x)  =  (1 - p_{D,k})\mathcal{D}_{k|k-1}(x) + \sum_{z\in Z_k} \mathcal{D}_{D,k}(x;z),
\label{eqn:PHDupdateki}
\end{split}
\end{equation}
\noindent where
\begin{equation}
\begin{split}
     \mathcal{D}_{D,k}(x;z)  =  \sum_{v = 1}^{V_{k|k-1}} w_{k}^{(v)}(z) \mathcal{N}(x; m_{k|k}^{(v)}(z), P_{k|k}^{(v)}), \nonumber
\label{eqn:PHDupdateDetki}
\end{split}
\end{equation}
\noindent
\begin{equation}
\begin{split}
     w^{(v)}_{k}(z)  =  \frac{p_{D,k} w^{(v)}_{k|k-1} q^{(v)}_{k}(z)}{c_{s_{k}}(z) +  p_{D,k} \sum_{l = 1}^{V_{k|k-1}} w^{(l)}_{k|k-1} q^{(l)}_{k}(z)}, \nonumber
\label{eqn:PHDupdatewwightki}
\end{split}
\end{equation}
\noindent
\begin{equation}
\begin{split}
     q^{(v)}_{k}(z)  =  \mathcal{N}(z; H_k m_{k|k-1}^{(v)}, R_k + H_k P_{k|k-1}^{(v)} H^T_k), \nonumber
\label{eqn:PHDupdateqki}
\end{split}
\end{equation}
\noindent
\begin{equation}
\begin{split}
     m^{(v)}_{k|k}(z)  =  m^{(v)}_{k|k-1} + K^{(v)}_k (z - H_k m_{k|k-1}^{(v)}), \nonumber
\label{eqn:PHDupdatemki}
\end{split}
\end{equation}
\noindent
\begin{equation}
\begin{split}
     P^{(v)}_{k|k}  =  [I -  K^{(v)}_k H_k] P_{k|k-1}^{(v)}, \nonumber
\label{eqn:PHDupdatepki}
\end{split}
\end{equation}
\noindent
\begin{equation}
\begin{split}
     K^{(v)}_k  =  P_{k|k-1}^{(v)} H^T_k [ H_k P_{k|k-1}^{(v)} H^T_k + R_k]^{-1}, \nonumber
\label{eqn:PHDupdateKki}
\end{split}
\end{equation}
\noindent The clutter intensity due to the scene, $c_{s_k}(z)$, in Eq.~(\ref{eqn:PHDupdateki}) is given by
\begin{equation}
    c_{s_k}(z) = \lambda_t c(z) = \lambda_{c} A c(z),
\label{eqn:Clutterscenei}
\end{equation}
\noindent where $c(.)$ is the uniform density over the surveillance region $A$, and $\lambda_{c}$ is the average number of clutter returns per unit volume i.e. $\lambda_t = \lambda_{c}A$. We set the clutter rate or false positive per image (fppi) $\lambda_t = 4$ in our experiment.

After update, weak Gaussian components with weight $w_k^{(v)} < T = 10^{-5}$ are pruned, and Gaussian components with Mahalanobis distance less than $U = 4$ pixels from each other are merged. These pruned and merged Gaussian components are predicted as existing targets in the next iteration. Finally, the Gaussian component of the posterior intensity with mean corresponding to the maximum weight is selected as a target state (position) estimate when the re-detection module is activated.

\subsection{Scale Estimation}

At the new estimated target position (or re-fined target position after re-detection in case of tracking failure), we construct an image pyramid for estimating its scale. Given a target size of $P \times Q$ in a test frame, we generate $S$ number of scale levels at the new estimated position i.e. for each $n \in \{\lfloor -\frac{S-1}{2}\rfloor,\lfloor -\frac{S-3}{2}\rfloor,...,\lfloor \frac{S-1}{2}\rfloor\}$, we extract an image patch $I_s$ of size $s P \times s Q$ centered at the new estimated target position, where scale $s = a^{n}$ and $a$ is the scale factor between the generated image pyramids. We uniformly resize all the generated image pyramids to $P \times Q$ again unlike~\cite{DanHagSha14}, and extracted HOG features particularly Felzenszwalb's variant~\cite{FelGirMcA10} to construct the scale feature pyramid. Then, the optimal scale $\hat{s}$ of a target at the estimated new position can be obtained by computing the correlation response maps $\hat{\mathbf{r}}_s$ of the scale correlation filter $\mathbf{w}_s$ to $I_s$ and find the scale at which the maximum response map can be obtained as

\begin{equation}
    \hat{s} = \underset{s}{\text{argmax}} \big(\hat{\mathbf{r}}_s \big),
\label{eqn:Scale}
\end{equation}
\noindent The scale correlation filter is updated using the new training sample at the estimated scale $I_{\hat{s}}$ by Eq.~(\ref{eqn:CorrelattionFilterModelUpdate}).

\begin{algorithm}[!htb]
\caption{Proposed tracking algorithm} \label{alg:ProposedTraAlg}
 \scriptsize
 \KwIn{Image $\mathbf{I}_{k}$, previous target position $\mathbf{x}_{k-1}$, previous correlation filters $\mathbf{w}^{(l)}_{t,k-1}$ and $\mathbf{w}_{s,k-1}$, previous SVM detector $D_r$}
 \KwOut{Estimated target position $\mathbf{x}_k = (x_k, y_k)$, updated correlation filters $\mathbf{w}^{(l)}_{t,k}$ and $\mathbf{w}_{s,k}$, updated SVM detector $D_r$}

 \Repeat{End of video sequences}{
  Crop out the searching window in frame k according to $(x_{k-1},y_{k-1})$ and extract multi-layer hybrid features and resize them to a fixed size\;

  \hrulefill \\
  \textit{// Translation estimation}\\
   \ForEach{layer l}{
     compute response map $\mathbf{r}^{(l)}$ using $\mathbf{w}^{(l)}_{t,k-1}$ and Eq.~(\ref{eqn:CorrelationFilterR});
   }

   Sum up the response maps of all layers element-wise according to their weight $\gamma(l)$ to get $\mathbf{r}(m,n)$ using Eq.~(\ref{eqn:CorrelationFilterRC});  \\
   Estimate the new target position $(x_k, y_k)$ by finding the maximum response of $\mathbf{r}(m,n)$ using Eq.~(\ref{eqn:EstimatedPosition});

\hrulefill \\
  \textit{// Apply GM-PHD filter} \\
  Update GM-PHD filter using the estimated target position $(x_k, y_k)$ as measurement but without re-fining it, just to update weight of GM-PHD filter for later use;

  \hrulefill \\
 \textit{// Target re-detection} \\
 \If{$\max \big(\mathbf{r}(m,n)\big) < T_{rd}$} {
        Use the detector $D_r$ to generate detection proposals $Z_k$ from high scores of incremental SVM;\\

        \hrulefill \\
  \textit{// Filtering using GM-PHD filter} \\
  Filter the generated candidate detections $Z_k$ using GM-PHD filter and select the detection with maximum weight as a re-detected target position $(x_{rk}, y_{rk})$. Then crop out the searching window at this re-detected position and compute its response map using Eq.~(\ref{eqn:CorrelationFilterR}) and Eq.~(\ref{eqn:CorrelationFilterRC}), and call it $\mathbf{r}_{rd}(m,n)$; \\
\If{$\max\big(\mathbf{r}_{rd}(m,n)\big) > T_{rd}$}  {
        $(x_k, y_k) = (x_{rk}, y_{rk})$ i.e. re-fine by the re-detected position; \\
}
}

\hrulefill \\
  \textit{// Scale estimation} \\
   Construct target image pyramid around $(x_k, y_k)$ and extract HOG features (resized to same size), and then compute the response maps $\hat{\mathbf{r}}_s$ using $\mathbf{w}_{s,k-1}$ and Eq.(\ref{eqn:CorrelationFilterR}), and then estimate its scale $\hat{s}_k$ using Eq.~(\ref{eqn:Scale}); \\

\hrulefill \\
 \textit{// Translation correlation model update} \\
  Crop out new patch centered at $(x_k, y_k)$ and extract multi-layer hybrid features and resize them to a fixed size; \\
\ForEach{layer l} {
   Update translation correlation filter $\mathbf{w}^{(l)}_{t,k}$ using Eq.~(\ref{eqn:CorrelattionFilterModelUpdate}); \\
}

%

\hrulefill \\
  \textit{// Scale correlation model update} \\
  Crop out new patch centered at $(x_k, y_k)$ with estimated scale $\hat{s}_k$ and extract HOG features and then update correlation filter $\mathbf{w}_{s,k}$ using Eq.~(\ref{eqn:CorrelattionFilterModelUpdate}); \\

\hrulefill \\
 \textit{// Update detector $D_r$} \\
\If{$\max \big(\mathbf{r}(m,n)\big) \geq T_{td}$} {
 Generate positive and negative samples around $(x_k, y_k)$ and then extract HOG, LUV color and normalized gradient magnitude features to train incremental SVM for updating its weight vector using Eq.~(\ref{eqn:KKTw}); \\

 }

}
\end{algorithm}
\normalsize

\section{Implementation Details} \label{Sec:Implementation}

\begin{figure*}[!htb]
\begin{center}
   \includegraphics[width=0.80\linewidth]{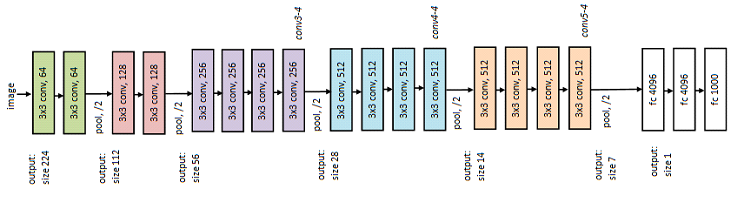} \\
\end{center}
   \caption{\small{VGG-Net 19~\cite{SimZis15}.}} 
\label{fig:VGGNet19}
\end{figure*}

The main steps of our proposed algorithm are presented in Algorithm~\ref{alg:ProposedTraAlg}. More implementation details with parameter settings are given as follows. For learning the translation correlation filter, we extract features from VGG-Net~\cite{SimZis15}, shown in Fig.~\ref{fig:VGGNet19}, trained on a large amount of object recognition data set (ImageNet)~\cite{DenDonSoc09} by first removing fully-connected layers. Particularly, we use outputs of \textit{conv}3-4, \textit{conv}4-4 and \textit{conv}5-4 convolutional layers as features ($l \in \{1, 2, 3\}$ and $d \in \{1, ...,D\}$), i.e. the outputs of rectilinear units (inputs of pooling) layers must be used to keep more spatial resolution. Hence, the CNN features we use has 3 layers ($L = 3$) and multiple channels ($D = 512$) for \textit{conv}5-4 and \textit{conv}4-4 layers and ($D = 256$) for \textit{conv}3-4 layer. For hand-crafted features, the HOG variant with 31 dimensions and color-naming with 11 dimensions are integrated to make 42 dimensional features which make a 4th layer in our hybrid multi-layer features. Given an image frame with a search window size of $\tilde{M} \times \tilde{N}$ which is about 2.8 times the target size to provide some context, we resize the multi-layer hybrid features to a fixed spatial size of $M \times N$ where $M = \frac{\tilde{M}}{4}$ and $N = \frac{\tilde{N}}{4}$. These hybrid features from each layer are weighted by a cosine window~\cite{HenCasMar14} to remove the boundary discontinuities, and then combined later on in Eq.~(\ref{eqn:CorrelationFilterRC}) for which we set $\gamma$ as 1, 0.4, 0.02 and 0.1 for the \textit{conv}5-4, \textit{conv}4-4, \textit{conv}3-4 and hand-crafted features, respectively. We set the regularization parameter of the ridge regression in Eq.~(\ref{eqn:RidgeRegression}) to $\lambda = 10^{-4}$, and a kernel bandwidth of the Gaussian function label in Eq.~(\ref{eqn:GaussianLabel}) to $\sigma = 0.1$. The learning rate for model update in Eq.~(\ref{eqn:CorrelattionFilterModelUpdate}) is set to $\eta = 0.01$.

For learning the scale correlation filter, we use the same parameter settings as above with some exceptions as follows. In this case we use HOG features~\cite{FelGirMcA10} with 31 bins i.e. it is treated as a single layer ($L=1$) but with multiple channels ($D = 31$). The number of scale spaces is set to $S = 31$ and the scale factor is set to $a = 1.04$. We use a linear kernel Eq.~(\ref{eqn:LinearKernel}) for learning both translation and scale correlation filters.

HOG, LUV color and normalized gradient magnitude features are used to train an incremental (online) SVM classifier for the re-detection module. For the objective function given in Eq.~(\ref{eqn:ObjectiveFn}), we use a Gaussian kernel, particularly for $Q_{ij} = y_i y_j K(\mathbf{x_i},\mathbf{x_j})$, and the regularization parameter $C$ is set to 2. Empirically, we set the activated re-detection threshold to $T_{rd} = 0.15$ and the trained detector threshold to $T_{td} = 0.40$. The parameters in Eq.~(\ref{eqn:label1}) are set as $\delta_p = 0.9$ and $\delta_n = 0.3$. For negative samples, we randomly sampled 3 times the number of positive samples satisfying $\delta_n = 0.3$ within the maximum search area of 4 times the target size. In the re-detection phase, we generated 5 high-score detection proposals from the trained online SVM around the estimated position within the maximum search area of 6 times the target size which were filtered using the GM-PHD filter to find the detection with the maximum weight removing the others as clutter. The implementation parameters are summarized in Table~\ref{tbl:parameters}.

\begin{table*}[!htb]
\begin{center}
\begin{tabular}{|l|c|c|c|c|c|c|c|c|c|c|c|c|r|}
\hline
Parameters & $\lambda$ &  $\sigma$ & $\eta$ & C & $T_{rd}$ & $T_{td}$ & $\delta_p$ & $\delta_n$ & $S$ & $a$ & $\lambda_t$ & $U$ & $T$ \\
\hline
Values  & $10^{-4}$ & 0.1 & 0.01 & 2 & 0.15 & 0.40 & 0.9 & 0.3 & 31 & 1.04 & 4 & 4 pixels & $10^{-5}$ \\
\hline
\end{tabular}
\end{center}
\caption{\small{Implementation parameters.}}
\label{tbl:parameters}
\end{table*}
\noindent

\noindent

\section{Experimental Results} \label{Sec:ExperimentalResults}

We evaluate our proposed tracking algorithm on both a large-scale online object tracking benchmark (OOTB)~\cite{WuLimYan13} and crowded scenes (medium and dense PETS 2009 data sets\footnote{http://www.cvg.reading.ac.uk/PETS2009/a.html}), and compared its performance with state-of-the-art trackers using the same parameter values for all the sequences. We quantitatively evaluate the robustness of the trackers using two metrics, precision and success rate based on center location error and bounding box overlap ratio, respectively, using one-pass evaluation (OPE) setting, running the trackers throughout a test sequence with initialization from the ground truth position in the first frame. The center location error computes the average Euclidean distance between the center locations of the tracked targets and the manually labeled ground truth positions of all the frames whereas bounding box overlap ratio computes the intersection over union of the tracked target and ground truth bounding boxes.

Our proposed tracking algorithm is implemented in MATLAB on a 3.0 GHz Intel Xeon CPU E5-1607 with 16 GB RAM.  We also use the MatConvNet toolbox~\cite{VedLen15} for CNN feature extraction where its forward propagation computation is transferred to a NVIDIA Quadro K5000, and our tracker runs at 5 fps on this setting. The re-detection step and forward propagation for feature extraction step are the main computational load steps of our tracking algorithm. We analyze our algorithm and then compare it with the state-of-the-art trackers both quantitatively and qualitatively on OOTB and PETS 2009 data sets separately as follows.

\subsection{Evaluation on OOTB}

OOTB~\cite{WuLimYan13} contains 50 fully annotated videos with substantial variations such as scale, occlusion, illumination, etc and is currently a popular tracking benchmark available in the computer vision community. In this experiment, we compare our proposed tracking algorithm with 6 state-of-the-art trackers including CF2~\cite{MaHunYan15}, LCT~\cite{MaYanZha15}, MEEM~\cite{ZhaMaScl14}, DLT~\cite{WanYeu13}, KCF~\cite{HenCasMar14} and SAMF~\cite{LiZhu15}, as well as 4 more top trackers included in the Benchmark~\cite{WuLimYan13}, particularly SCM~\cite{ZhoLuYan12}, ASLA~\cite{JiaLuYan12}, TLD~\cite{ZdeKryJir12} and Struck~\cite{HarSafTor11} both quantitatively and qualitatively.

\begin{figure}
\centering
\begin{minipage}[c]{0.48\textwidth}
\centering
    \includegraphics[width=0.70\linewidth]{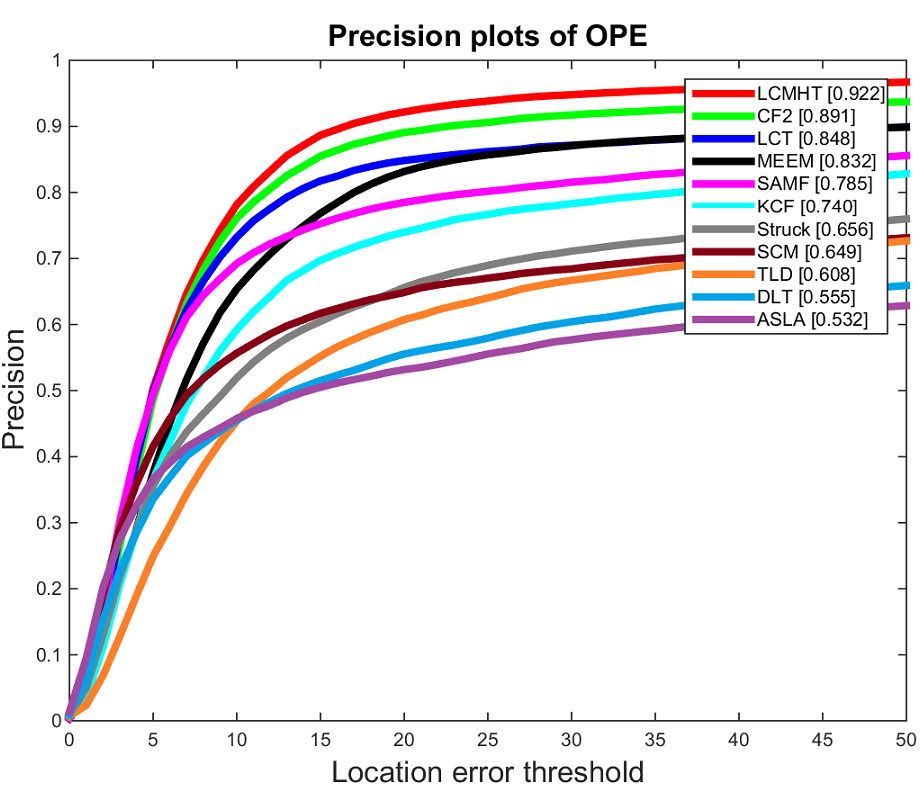}
    \includegraphics[width=0.70\linewidth]{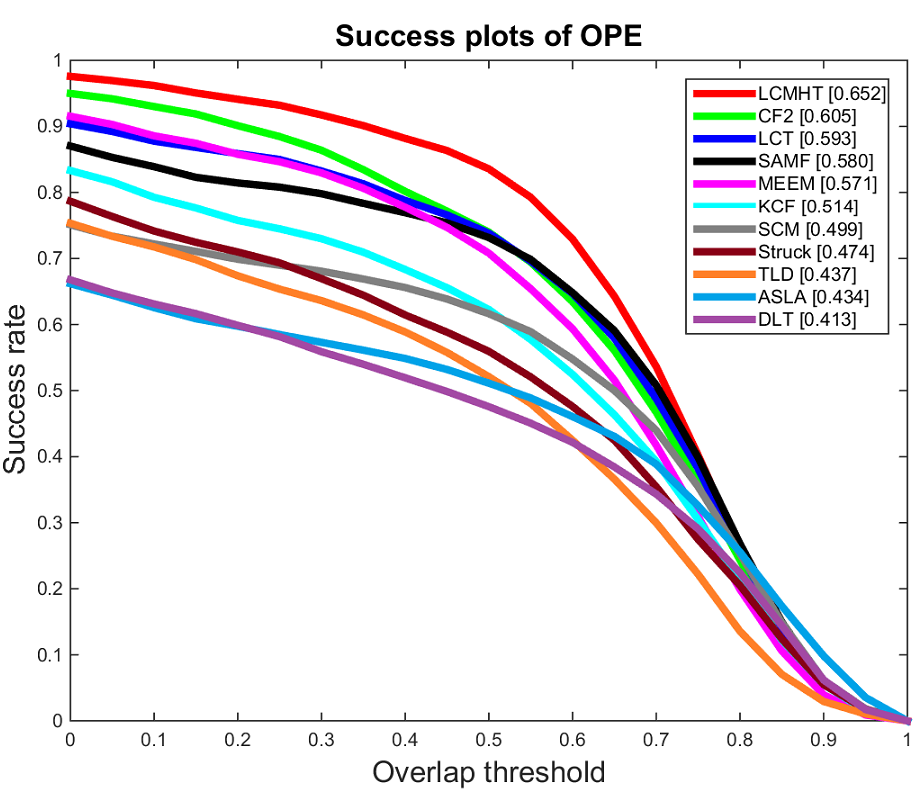}
    \caption{\small{Distance precision (top) and overlap success (bottom) plots on OOTB using one-pass evaluation (OPE). The legend for distance precision contains threshold scores at 20 pixels while the legend for overlap success contains the AUC score of each tracker; the larger, the better.}}
    \label{fig:OPEootb}
\end{minipage}
\end{figure}

\textbf{Quantitative Evaluation}: We evaluate our proposed tracking algorithm quantitatively and compare with other algorithms as summarized in Fig.~\ref{fig:OPEootb} using precision plots (left) and success plots (right) based on center location error and bounding box overlap ratio, respectively. Our proposed tracking algorithm, denoted by LCMHT, outperforms the state-of-the-art trackers in both precision and success measures by rankings given in the legends using a distance precision of threshold scores at 20 pixels and overlap success of area-under-curve (AUC) score for each tracker, respectively. This is because a hybrid of multi-layer CNN, HOG and color-naming features is more effective to represent the target than their individual features separately i.e. our proposed tracking algorithm integrates a hybrid of multi-layer CNN and traditional (HOG and color-naming) features for learning a translation correlation filter, and uses the GM-PHD filter for temporally filtering generated high score detection proposals during a re-detection phase for removing clutter so that it can re-detect the target even in a cluttered environment.

\begin{figure*} [!htb] 
  \begin{center}
  {\label{fig:VisCardinality} \includegraphics[height=0.210\textwidth]{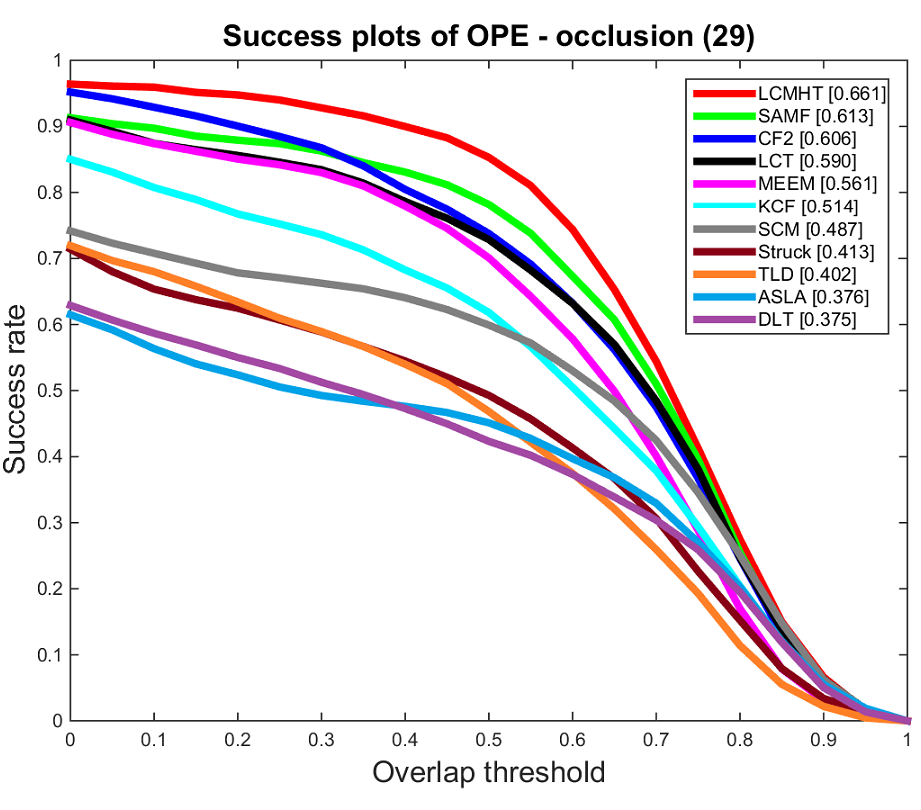}} 
  {\label{fig:VisOSPAerror} \includegraphics[height=0.210\textwidth]{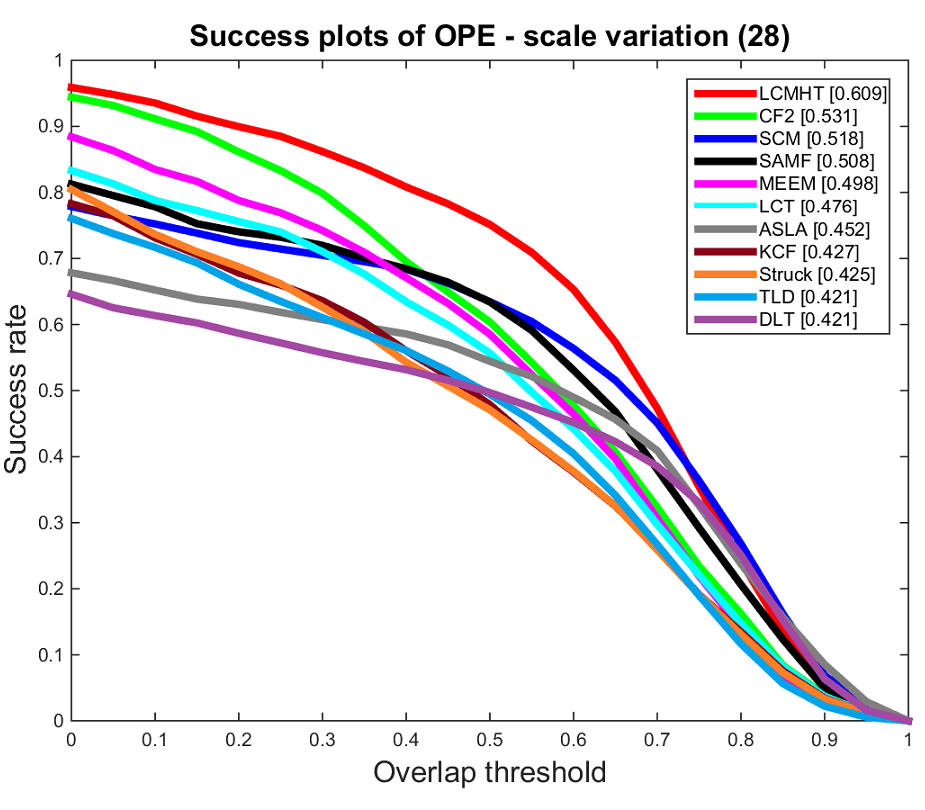}}
   {\label{fig:VisOSPAerror} \includegraphics[height=0.210\textwidth]{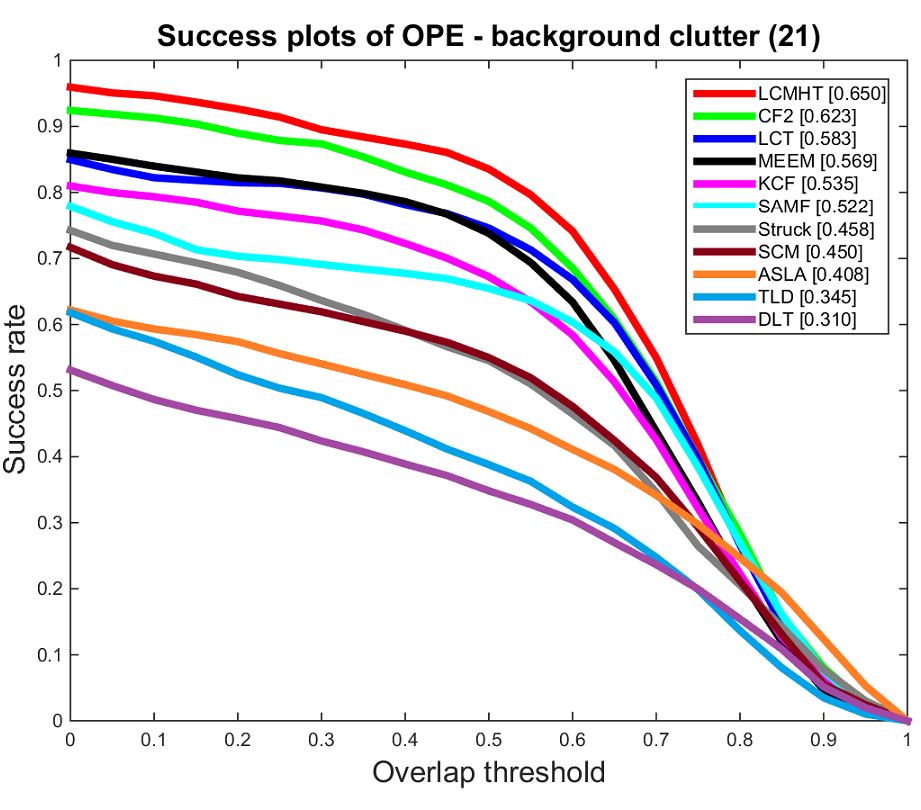}}
  {\label{fig:VisOSPAerror} \includegraphics[height=0.210\textwidth]{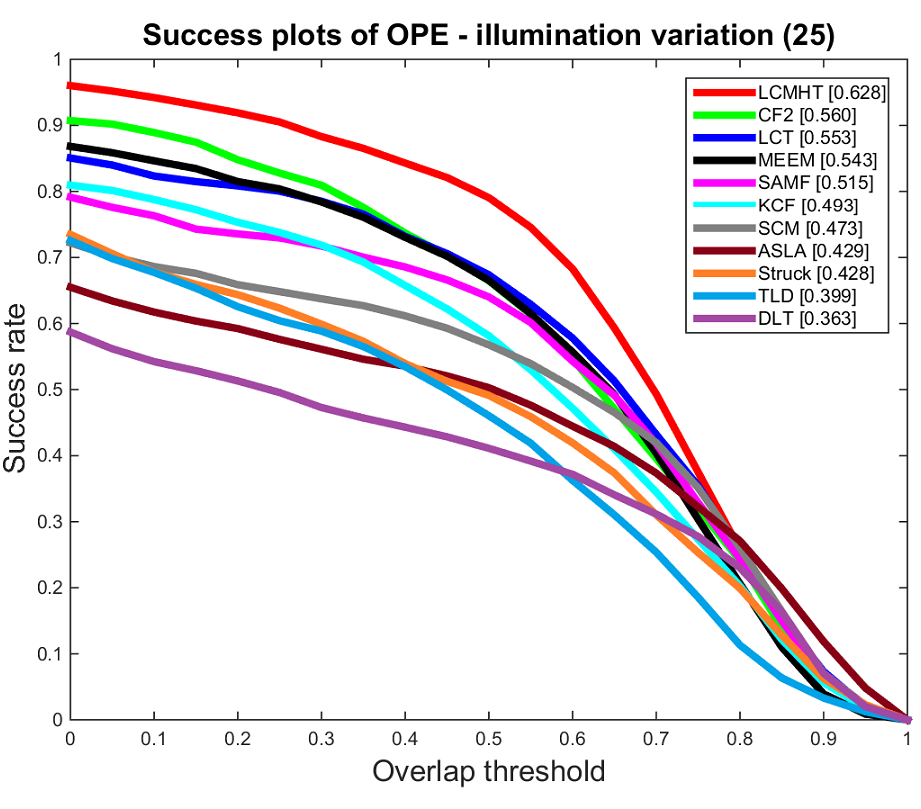}} \\
  {\label{fig:VisOSPAerror} \includegraphics[height=0.210\textwidth]{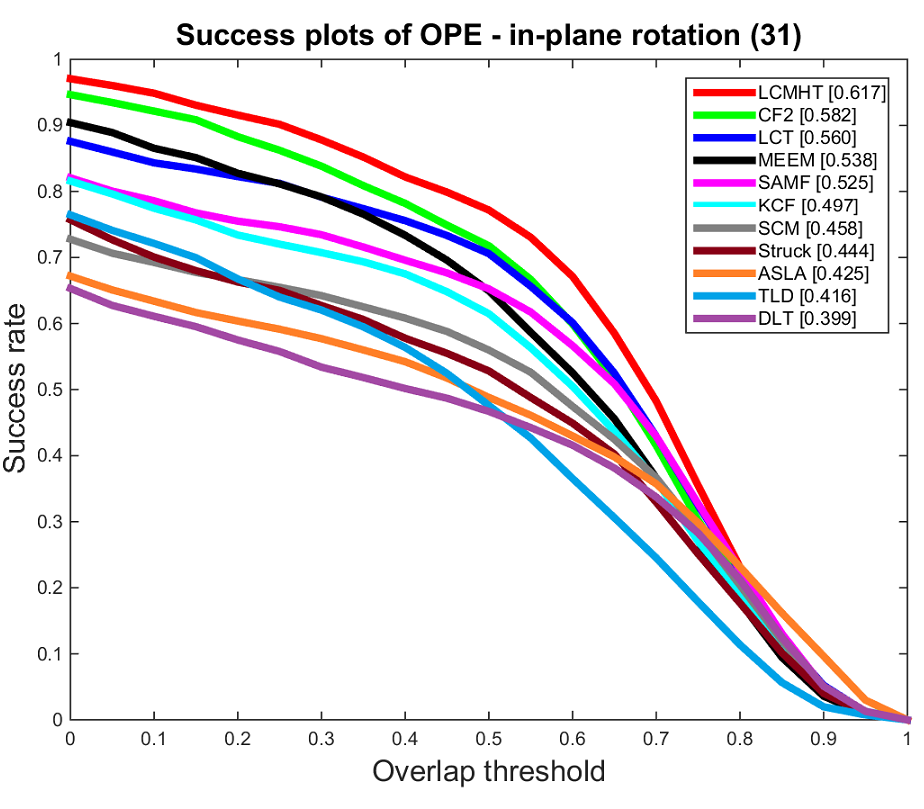}}
  {\label{fig:VisOSPAerror} \includegraphics[height=0.210\textwidth]{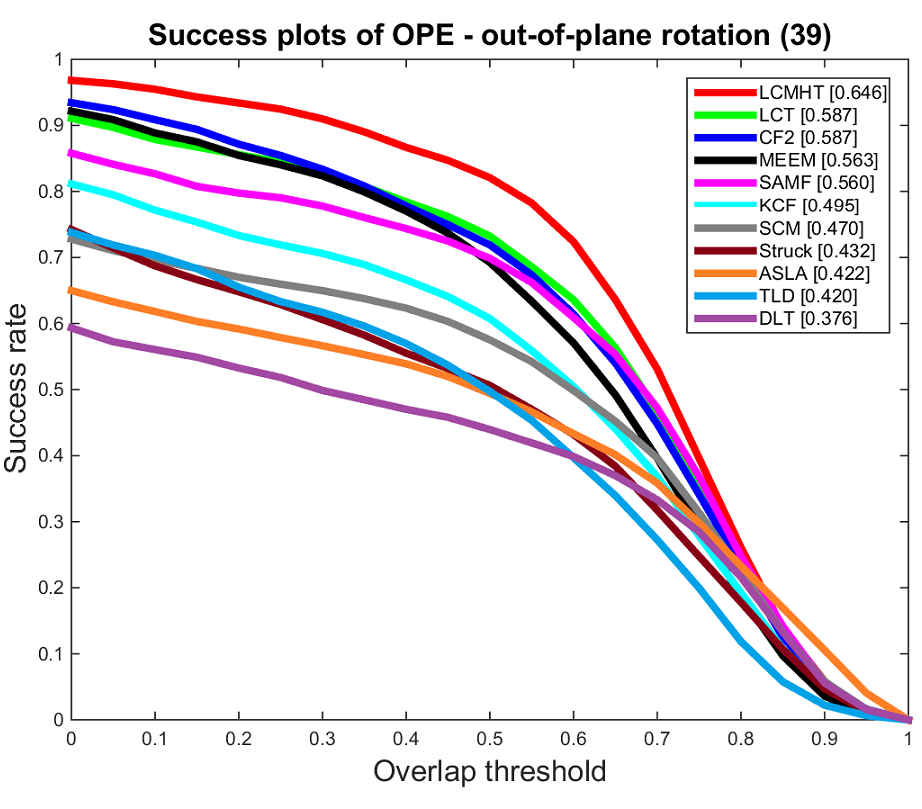}}
  {\label{fig:VisOSPAerror} \includegraphics[height=0.210\textwidth]{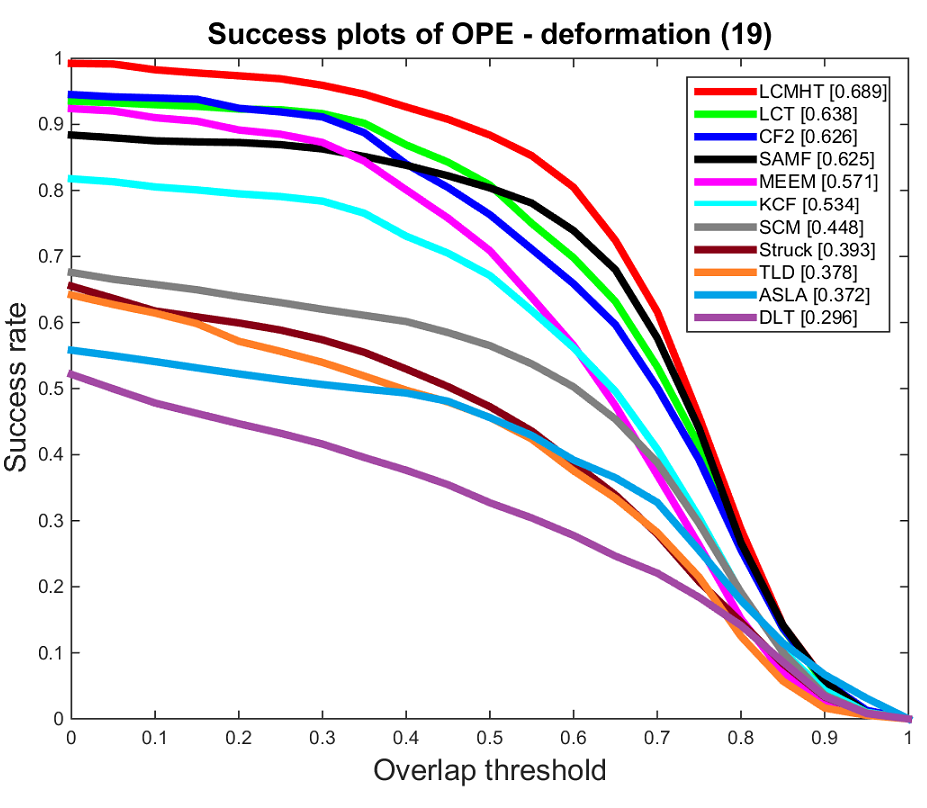}}
   {\label{fig:VisOSPAerror} \includegraphics[height=0.210\textwidth]{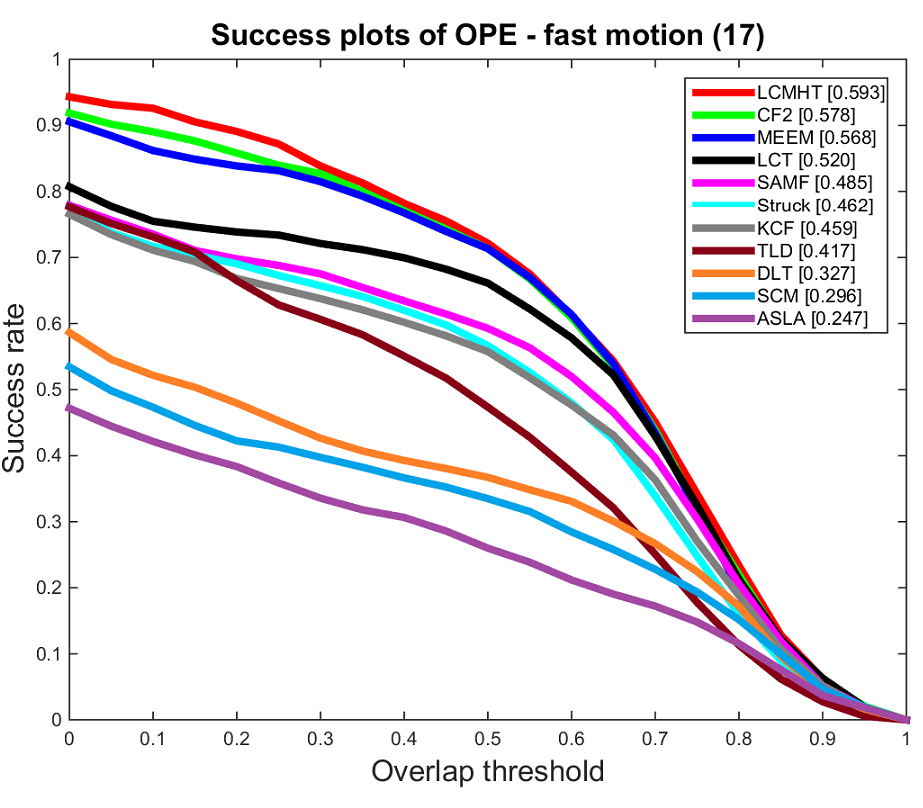}}
  \end{center}
   \caption{\small{Success plots on OOTB using one-pass evaluation (OPE) for 8 challenge attributes: occlusion, scale variation, background clutter, illumination variation, in-plane rotation, out-of-plane rotation, deformation, and fast motion. The legend contains the AUC score of each tracker; the larger, the better. }}  
  \label{fig:OPEootbAttributes}
\end{figure*}

\textbf{Attribute-based Evaluation}: For the detailed performance analysis of each of the trackers, we also report the results on various challenge attributes in OOTB~\cite{WuLimYan13} such as occlusion, scale variation, illumination variation, etc. As shown in Fig.~\ref{fig:OPEootbAttributes}, our proposed tracker outperforms the state-of-the-art trackers in almost all challenge attributes. In particular, our proposed tracker (LCMHT) performs significantly better than all trackers on the occlusion attribute since it includes a re-detection module which can re-acquire the target in case the tracker fails even in cluttered environments by removing clutter using GM-PHD filter. Similarly, our tracker also outperforms other trackers on the scale variation attribute since our tracker elegantly estimates the scale of the tracker at the newly estimated target positions. The LCT algorithm includes both re-detection and scale estimation modules, however, our proposed tracker still outperforms the LCT algorithm by a large margin as shown in Fig.~\ref{fig:OPEootbAttributes} since our tracker uses better visual features for translation estimation and re-detection. Furthermore, our proposed algorithm applies scale estimation after translation and re-detection steps (if activated) rather than only after the translation estimation step as in the LCT algorithm, though both methods use similar visual features (HOG) to learn the scale correlation filter.

\noindent

\begin{figure*}[!htb]
\begin{center}
   \includegraphics[width=1.0\linewidth]{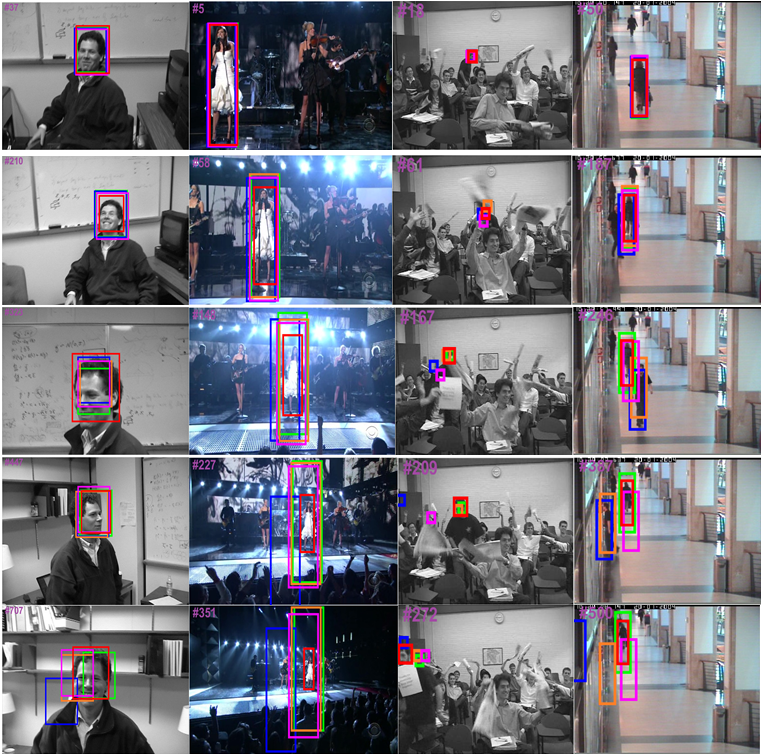} \\ 
   \includegraphics[width=0.60\linewidth]{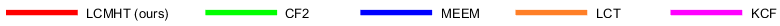}
\end{center}
   \caption{\small{Qualitative results of our proposed LCMHT algorithm, CF2~\cite{MaHunYan15}, MEEM~\cite{ZhaMaScl14}, LCT~\cite{MaYanZha15} and KCF~\cite{HenCasMar14} on some challenging sequences of OOTB (Fleetface, Singer1, Freeman4, and Walking2 from left column to right column, respectively).}} 
\label{fig:qualitativeOOTB}
\end{figure*}

\textbf{Qualitative Evaluation}: We compare our proposed tracking algorithm (LCMHT) with four other state-of-the-art trackers namely CF2~\cite{MaHunYan15}, MEEM~\cite{ZhaMaScl14}, LCT~\cite{MaYanZha15} and KCF~\cite{HenCasMar14} on some challenging sequences of OOTB qualitatively as shown in Fig.~\ref{fig:qualitativeOOTB}. CF2 uses hierarchical CNN features but is not as effective as our tracker which combines hierarchical CNN features with HOG and color-naming traditional features as can be observed on the sequence \textit{Fleetface} (first column on Fig.~\ref{fig:qualitativeOOTB}). LCT and KCF also use correlation filters using traditional features but still they are not as accurate as our tracker. MEEM uses many classifiers together to re-initialize the tracker in case of tracking failures but it can not re-detect the target on this sequence. Similarly, it can not re-detect the target on sequences \textit{Singer1} (second column), \textit{Freeman4} (third column) and \textit{Walking2} (forth column) as well. LCT includes re-detection and scale estimation components, however, it can not handle large scale changes as in sequence \textit{Singer1} (second column), and it can not re-initialize the tracker as in sequence \textit{Walking2} (forth column). More importantly, the sequence \textit{Freeman4} undergoes not only heavy occlusion in a cluttered environment but also scale variation, in-plane and out-of-plane rotations. The LCT algorithm which is equipped with both re-detection and scale estimation modules is not effective on this sequence like the other algorithms. However, only our proposed tracker tracks the target till the end of the sequence not only handling the scale change but also re-detecting the target when it fails. This sequence is a typical example which is related to our next evaluation on PETS 2009 data sets on which our proposed algorithm outperforms the other trackers by a large margin.

\noindent

\subsection{Evaluation on PETS 2009 Data Sets}

We label the upper part (head + neck) of representative targets in both medium and dense PETS 2009 data sets to analyze our proposed tracking algorithm. In this experiment, our goal is to analyze our proposed tracking algorithm and other available state-of-the-art tracking algorithms to see whether they can successfully be applied for tracking a target of interest in occluded and cluttered environments. Accordingly, we compare our proposed tracking algorithm with 6 state-of-the-art trackers including CF2~\cite{MaHunYan15}, LCT~\cite{MaYanZha15}, MEEM~\cite{ZhaMaScl14}, DSST~\cite{DanHagSha14}, KCF~\cite{HenCasMar14} and SAMF~\cite{LiZhu15}, as well as 4 more top trackers included in the Benchmark~\cite{WuLimYan13}, particularly SCM~\cite{ZhoLuYan12}, ASLA~\cite{JiaLuYan12}, CSK~\cite{HenCasMar12} and IVT~\cite{RosLimLin08} both quantitatively and qualitatively.

\begin{figure}
\centering
\begin{minipage}[c]{0.48\textwidth}
\centering
    \includegraphics[width=0.70\linewidth]{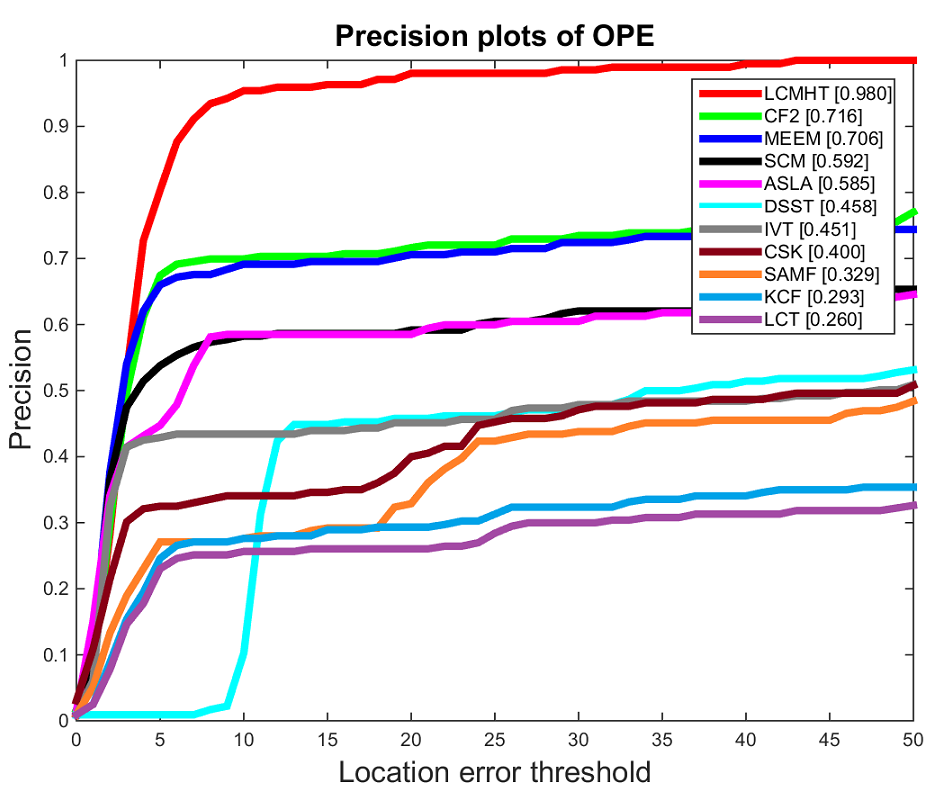}
    \includegraphics[width=0.70\linewidth]{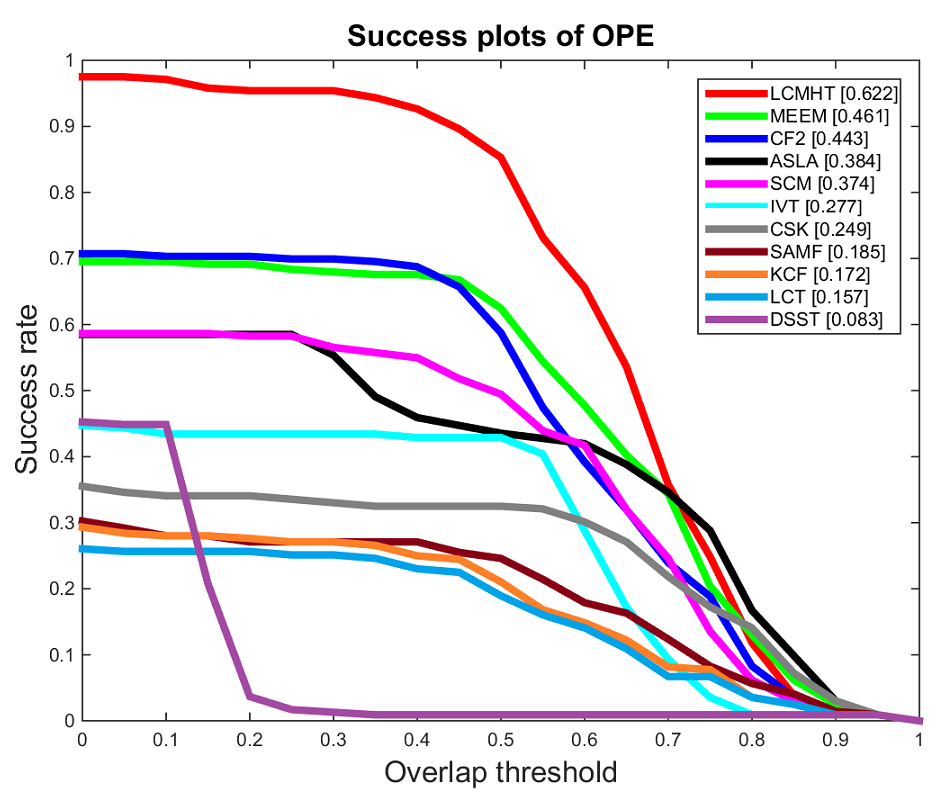}
    \caption{\small{Distance precision (top) and overlap success (bottom) plots on PETS data sets using one-pass evaluation (OPE). The legend for distance precision contains threshold scores at 20 pixels while the legend for overlap success contains the AUC score of each tracker; the larger, the better.}}
    \label{fig:OPEpets}
\end{minipage}
\end{figure}

\textbf{Quantitative Evaluation}: The evaluation results of precision plots (left) and success plots (right) based on center location error and bounding box overlap ratio, respectively, are shown in Fig.~\ref{fig:OPEpets}. Our proposed tracking algorithm, denoted by LCMHT, outperforms the state-of-the-art trackers by a large margin on PETS 2009 data sets in both precision and success rate measures. The rankings are given in distance precision of threshold scores at 20 pixels and overlap success of AUC score for each tracker as given in the legends.

The second and third ranked trackers are CF2~\cite{MaHunYan15} and MEEM~\cite{ZhaMaScl14} for precision plots, respectively, and viceversa for success plots on PETS 2009 data sets. However, on OOTB, CF2 outperforms MEEM significantly being second to our proposed tracking algorithm. The most important thing to give attention is on the performance of LCT~\cite{MaYanZha15}. This algorithm is ranked third on the OOTB as shown in Fig.~\ref{fig:OPEootb}, however, it performs least well on the precision plots and second from last on success plots on PETS 2009 data sets. Surprisingly, this algorithm was developed by learning three different discriminative correlation filters and even included a re-detection module for long-term tracking problems. Though it performs reasonably on the OOTB, its performance on occluded and cluttered environments such as PETS 2009 data sets is poor due to using less robust visual features in such environments. Even CF2 which uses CNN features has low performance compared to our proposed algorithm on the PETS 2009 data sets. Since our proposed tracking algorithm integrates a hybrid of multi-layer CNN and traditional features for learning the translation correlation filter and GM-PHD filter for temporally filtering generated high score detection proposals during a re-detection phase for removing clutter, it outperforms all the available trackers significantly. This closes the model-free tracking research gap between sparse and crowded environments.

\begin{figure*}[!htb]
\begin{center}
   \includegraphics[width=0.85\linewidth]{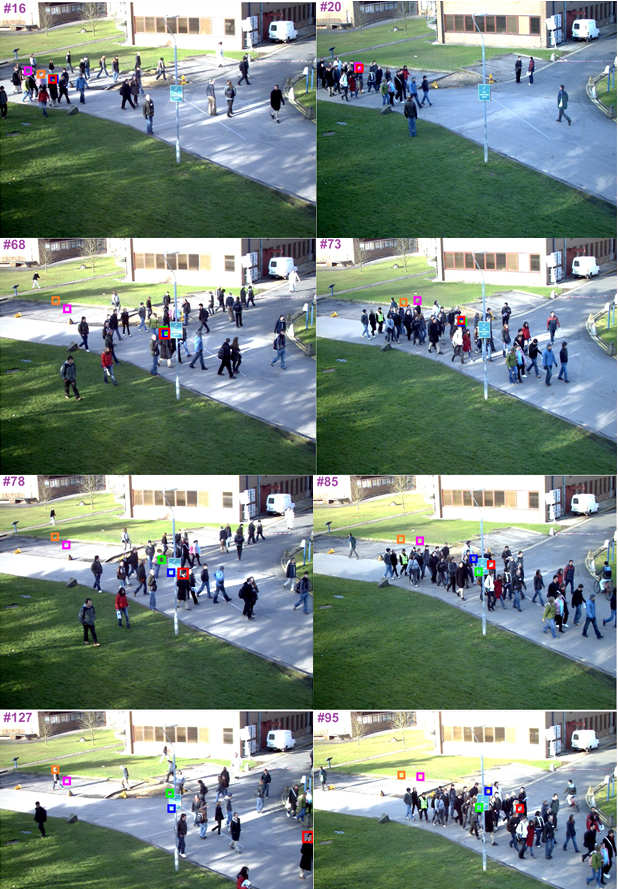} \\ 
   \includegraphics[width=0.60\linewidth]{OOTBlogo22.png}
\end{center}
   \caption{\small{Qualitative results of our proposed algorithm LCMHT, CF2~\cite{MaHunYan15}, MEEM~\cite{ZhaMaScl14}, LCT~\cite{MaYanZha15} and KCF~\cite{HenCasMar14} on PETS 2009 medium density (left column) and dense (right column) data sets.}} 
\label{fig:qualitativePETS}
\end{figure*}

\begin{figure*}[!htb]
\begin{center}
   \includegraphics[width=0.495\linewidth]{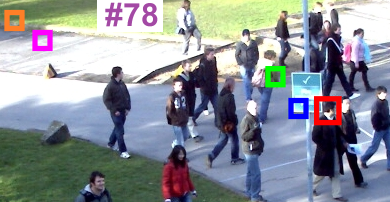}
   \includegraphics[width=0.495\linewidth]{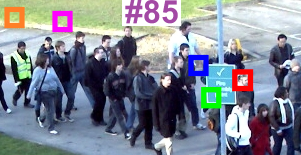} \\
   \includegraphics[width=0.60\linewidth]{OOTBlogo22.png}
\end{center}
   \caption{\small{Qualitative results of our proposed LCMHT algorithm, CF2~\cite{MaHunYan15}, MEEM~\cite{ZhaMaScl14}, LCT~\cite{MaYanZha15} and KCF~\cite{HenCasMar14} on PETS 2009 medium density (left, frame 78) and dense (right, frame 85) data sets, just after occlusion by cropping and enlarging.}} 
\label{fig:qualitativePETSocclusion}
\end{figure*}

\textbf{Qualitative Evaluation}: Fig.~\ref{fig:qualitativePETS} presents the performance of our proposed tracker qualitatively compared to the state-of-the-art trackers. In this case, we show the comparison of four representative trackers to our proposed algorithm: CF2~\cite{MaHunYan15}, MEEM~\cite{ZhaMaScl14}, LCT~\cite{MaYanZha15}, and KCF~\cite{HenCasMar14} as shown in Fig~\ref{fig:qualitativePETS}. On the medium density PETS 2009 data set (left column), LCT and KCF lose the target even on the first 16 frames. Though, the CF2 and MEEM trackers track the target well, they could not re-detect the target after the occlusion i.e. only our proposed tracking algorithm tracks the target till the end of the sequence by re-initializing the tracker after the occlusion. We show the cropped and enlarged re-detection just after occlusion in Fig.~\ref{fig:qualitativePETSocclusion}. On the dense PETS data set (right column),  all trackers track the target on the first 20 frames but LCT and KCF lose the target before 73 frames. Similar to the medium density PETS data set, the CF2 and MEEM trackers track the target before they lose it due to occlusion. Only our proposed tracking algorithm, LCMHT, re-detects the target and tracks it till the end of the sequence in such dense environments due to two reasons. First, it incorporates both lower and higher CNN layers in combination with traditional features (HOG and color-naming) in a multi-layer to learn the translation correlation filter that is robust to appearance variations of targets. Second, it includes a re-detection module which generates high score detection proposals during a re-detection phase and then filter them using GM-PHD filter to remove clutter due to background and other uninterested targets so that it can re-detect the target in such cluttered and dense environment. These make our proposed tracking algorithm outperform the other state-of-the-art trackers.

\noindent

\noindent

\section{Conclusions} \label{Sec:Conclusion}

We have developed a novel long-term visual tracking algorithm by learning discriminative correlation filters and an incremental SVM classifier that can be applied for tracking of a target of interest in sparse as well as in crowded environments. We learn two different discriminative correlation filters: translation and scale correlation filters. For the translation correlation filter, we combine a hybrid of multi-layer CNN features trained on a large amount of object recognition data set (ImageNet) and traditional (HOG and color-naming) features in proper proportion. For the CNN part, we combine the advantages of both lower and higher convolutional layers to capture spatial details for precise localization and semantic information for handling appearance variations, respectively. We also include a re-detection module using HOG, LUV color and normalized gradient magnitude features for re-initializing the tracker in case of tracking failures due to long-term occlusions by training an incremental SVM from the most confident frames. The re-detection module generates high score detection proposals which are temporally filtered using a GM-PHD filter for removing clutter. The Gaussian component with maximum weight is selected as a state estimate which re-fines the object location when a re-detection module is activated. For the scale correlation filter, we use HOG features to construct a target pyramid around the estimated or re-detected position for estimating the scale of the target. Extensive experiments on both OOTB and PETS 2009 data sets show that our proposed algorithm significantly outperforms state-of-the-art trackers by 3.48\% in distance precision and 7.77\% in overlap success on sparse (OOTB) data sets, and by 36.87\% in distance precision and 34.92\% in overlap success on dense (PETS 2009) data sets. We conclude that learning correlation filters using an appropriate combination of CNN and traditional features as well as including a re-detection module using incremental SVM and GM-PHD filter can give better results than many existing approaches.

\section*{Acknowledgment}

We would like to acknowledge the support of the Engineering and Physical Sciences Research Council (EPSRC), grant references EP/K009931, EP/J015180 and a James Watt Scholarship.

\bibliographystyle{IEEEtran}
\bibliography{egbib}




\end{document}